\documentclass{article}

\usepackage[round]{natbib}

\usepackage[margin=2.5cm]{geometry}
\usepackage[utf8]{inputenc} 
\usepackage[T1]{fontenc}    
\usepackage{hyperref}       
\usepackage{url}
\usepackage{booktabs}       
\usepackage{amsfonts}       
\usepackage{nicefrac}       
\usepackage{microtype}      
\usepackage{siunitx}
\usepackage{soul}
\usepackage{subcaption}
\usepackage{fontawesome5}
\usepackage{graphicx}
\usepackage{float}

\usepackage{amssymb}
\usepackage{pifont}

\usepackage{titling}

\usepackage[dvipsnames]{xcolor}
\usepackage{amsmath}
\usepackage{enumitem}

\newcommand{\cmark}{\textcolor{Green}{\ding{51}}}%
\newcommand{\xmark}{\textcolor{Red}{\ding{55}}}%

\newcommand{\hltodo}[1]{}

\newcommand{\ie}{\textit{i.e.}}
\newcommand{\eg}{\textit{e.g.}}

\title{When are radiology reports useful for training medical image classifiers?}

\author{%
  Herman Bergström\thanks{Corresponding author: \texttt{hermanb@chalmers.se}}$^{\ 1}$, Zhongqi Yue$^1$, and
  Fredrik D. Johansson$^1$ \\
  \small{$^1$Department of Computer Science \& Engineering,} \\ 
  \small{Chalmers University of Technology and University of Gothenburg}
}

\date{}

\begin{document}

\maketitle
\begin{abstract}
\noindent Medical images used to train machine learning models are often accompanied by radiology reports containing rich expert annotations.
However, relying on these reports as inputs for clinical prediction requires the timely manual work of a trained radiologist. 
This raises a natural question: when can radiology reports be leveraged \emph{during training} to improve image-only classification?
Prior works are limited to evaluating pre-trained image representations by fine-tuning them to predict diagnostic labels, often extracted from reports, ignoring tasks with labels that are weakly associated with the text.
To address this gap, we conduct a systematic study of how radiology reports can be used during both pre-training and fine-tuning, across diagnostic and prognostic tasks (e.g., 12-month readmission), and under varying training set sizes.
Our findings reveal that: (1) Leveraging reports during pre-training is beneficial for downstream classification tasks where the label is well-represented in the text; however, pre-training through explicit image-text alignment can be detrimental in settings where it's not; (2) Fine-tuning with reports can lead to significant improvements and even have a larger impact than the pre-training method in certain settings. 
These results provide actionable insights into when and how to leverage privileged text data to train medical image classifiers while highlighting gaps in current research.

\end{abstract}
%
%
\section{Introduction}
\label{sec:1}

Radiology reports containing key findings from medical images are routinely produced in clinical practice. These texts have recently received much attention within the machine learning literature \citep{bannur2024maira, yang2023radiology, tanno2025collaboration}, and have been shown to be predictive of various patient outcomes, such as hospital readmission \citep{huang2019clinicalbert}, ICU mortality \citep{lin2021empirical}, and cancer progression \citep{batch2022developing}. Requiring manually written reports for test-time predictions is undesirable if the same prediction could be made from the medical image itself, but the prevalence of reports in retrospective data makes them attractive to use during training. In particular, reports can be used to improve few-shot or small-sample training of medical image classifiers, either by i) incorporating them in a \emph{pre-training} objective \citep{zhang_contrastive_2022}, or ii) viewing them as privileged information (PI) when \emph{fine-tuning} for a specific task \citep{vapnik2009new}.

\begin{figure*}[h]
    \centering
    \footnotesize
    \includegraphics[width=0.90\linewidth]{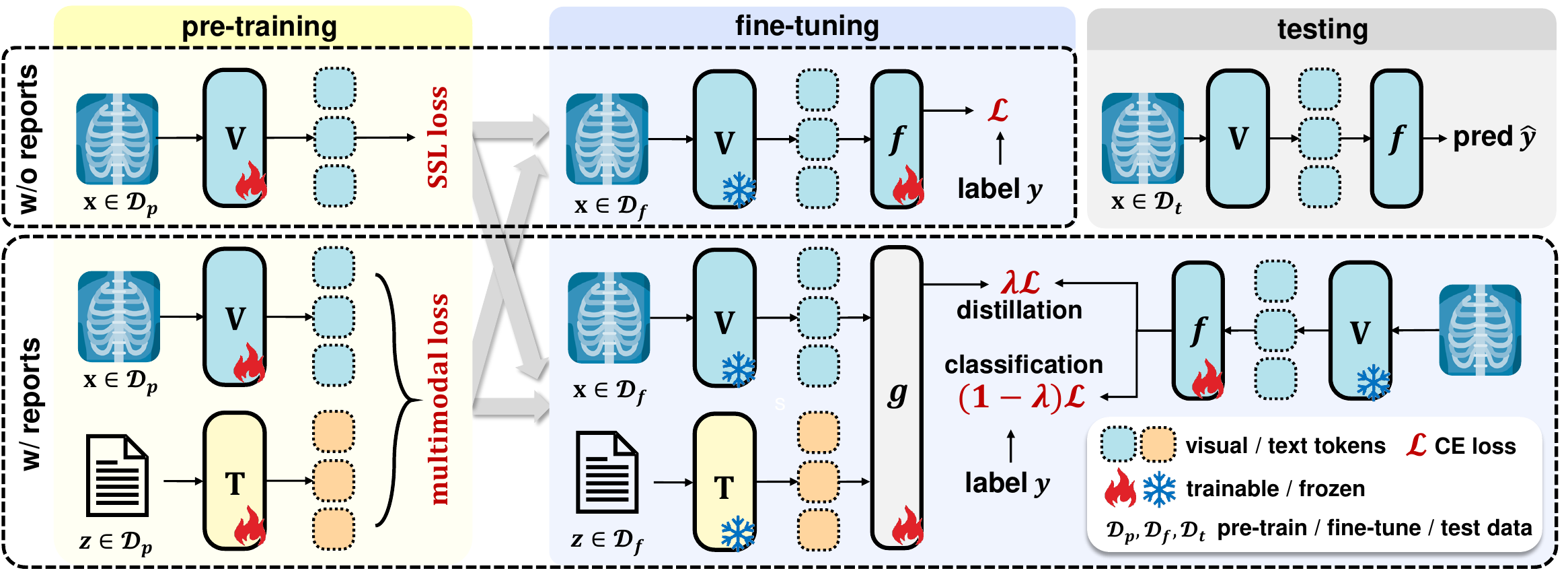}
    \caption{Training a medical image classifier with or without reports, divided into: pre-training and fine-tuning. Note that the classifier input is only a medical image at test time. ${V}$: visual encoder; ${T}$: text encoder; $f$: classifier; $g$: a privileged classifier using both image and report for prediction.}
    \label{fig:setup}
    \vspace{-4mm}
\end{figure*}

Prior work has almost exclusively focused on the potential role of reports in \emph{pre-training} image encoders---either incorporating them as paired supervision through multimodal objectives~\citep{huang_gloria_2021, zhou2023advancing}, or omitting them in favor of self-supervised learning~\citep{perez2024rad, zhou2021models}---with evaluation performed via report-free fine-tuning on \emph{diagnostic} classification tasks, where labels are often extracted from the reports themselves~\citep{johnson2019mimic,irvin2019chexpert, bustos2020padchest}.
As a result, the roles of radiology reports in \emph{fine-tuning} and their potential utility for tasks beyond diagnosis remain largely unexplored.
For instance, pre-training with reports instead of relying on image-only self-supervision has been shown to yield representations that improve performance in diagnostic tasks~\citep{zhang_contrastive_2022, huang_gloria_2021}.
Yet it is unclear whether these results generalize to \emph{prognostic} tasks---which aim to predict future clinical events (\eg, 12-month readmission or 3-day discharge) that may not be explicitly captured in the original report---and intriguingly, whether incorporating reports during fine-tuning can compensate for the potential limitations of a self-supervised pre-trained representation or a small fine-tuning data set.

To address these gaps, we compile a benchmark that systematically investigates the usefulness of radiology reports for training medical image classifiers during both pre-training and fine-tuning. We summarize our paradigm in Figure~\ref{fig:setup}.
In the pre-training stage, we study six popular models that differ in both architecture (ResNet~\citep{he2016deep} vs. ViT~\citep{dosovitskiy_image_2021}), and learning objective, including self-supervised learning~\citep{perez2024rad, zhou2020comparing, xiao2023delving}, CLIP-style text-image alignment~\citep{huang_gloria_2021,bannur_learning_2023,zhang2023biomedclip}, and masked image-text modeling~\citep{zhou2023advancing}.
For fine-tuning, we begin by highlighting that the relationships among the medical image (radiograph) $X$, report $Z$ and label $Y$ can vary significantly across diagnostic, prognostic, and other tasks, which in turn affects the utility of $Z$ when training a classifier to predict $Y$ from $X$.
Motivated by this, we study a range of classification tasks with varying correlation and causality between $Z$ and $Y$, and assess how using reports during fine-tuning by general distillation~\citep{lopez-paz_unifying_2016} impacts performance in each setting.
To the best of our knowledge, this is the first study of distillation from radiology reports, offering new insights into how privileged textual information can be transferred to image-only classifiers.

Our study leads to several interesting and novel findings:
(1) Pre-training with report supervision is beneficial for diagnostic tasks at moderate sample sizes when the label is \emph{strongly} correlated with the text.
(2) Explicitly aligning image and text embeddings hurts downstream performance when the label is not captured well by the report, something that is more prominent for non-diagnostic prediction tasks. Crucially, we find that methods relying on text supervision \emph{in addition to} self-supervision avoid this pitfall.
(3) Incorporating reports during fine-tuning, even when they are not available at test time, can yield substantial gains in accuracy.
\section{Method}
\label{sec:3}

Our goal is to study when training using radiology reports helps solve a $C$-way medical image-only classification task at test time, \ie, predicting the label $y\in\{1,\ldots,C\}$ from \emph{only} a medical image $\mathbf{x}$.
We consider each training sample $i$ in a data set $\mathcal{D}$ as a triplet $(X=\mathbf{x}_i, Z=z_i, Y=y_i)$, where the image-label pair $(\mathbf{x}_i,y_i)$ is additionally annotated with a radiology report $z_i$ derived from $\mathbf{x}_i$ to describe clinically significant observations. For brevity, we omit the subscript $i$ when it is unnecessary to highlight the sample index.
Training is divided into pre-training and fine-tuning. Next, we describe training procedures for both stages, with or without $Z$.

\subsection{Pre-training with and without reports}
\label{sec:3.1}

\begin{figure*}[t]
    \centering
    \includegraphics[width=0.9\linewidth]{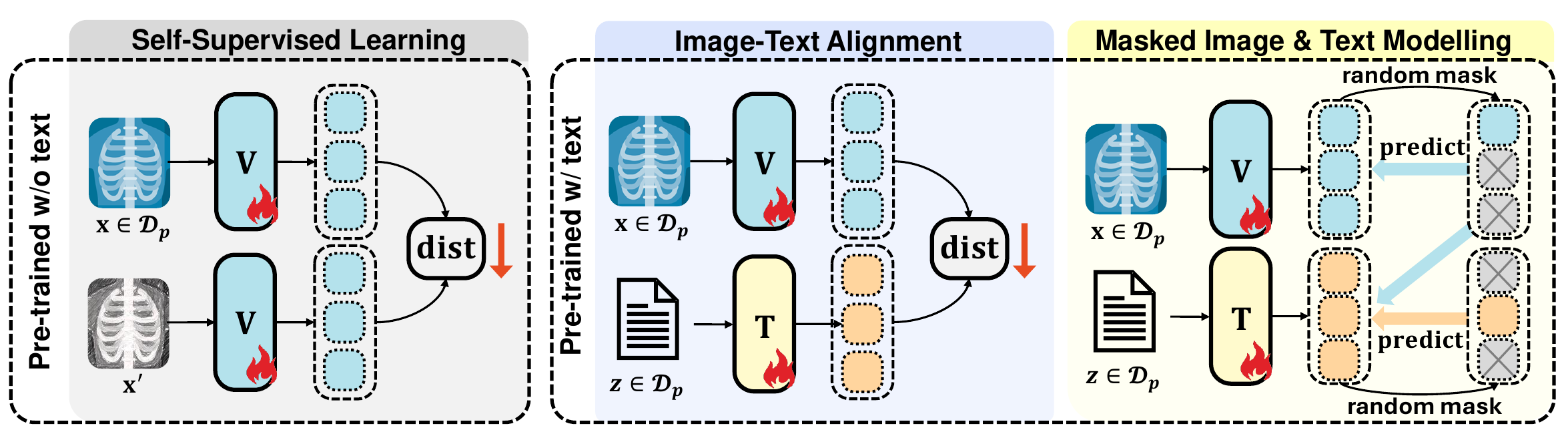}
    \caption{Three categories of pre-training objectives considered in this work, where $\mathbf{x}'$ is an augmented view of $\mathbf{x}$, $\mathcal{D}_p$ is the pre-training dataset, the fire symbol indicates trainable models, and $\mathrm{dist}$ denotes a distance function. $V$ and $T$ represent image (visual) and text encoders, respectively.}
    \label{fig:2}
\end{figure*}

In pre-training, we aim to learn a representation of images $\mathbf{x}$ through a visual encoder $V$, which maps each $\mathbf{x}$ into a sequence of visual tokens, \eg, the patch embeddings in ViT~\citep{dosovitskiy_image_2021} or convolutional feature maps in ResNet~\citep{he2016deep}. When medical reports are used in pre-training, analogously, we use a text encoder $T$, such as a BERT-style transformer, which maps each ${z}$ into a sequence of text tokens.
We consider 7 different models pre-trained for medical images from the literature, summarized in Table~\ref{tab:backbone}, illustrated in Figure~\ref{fig:2}, and described below.
\paragraph{Self-Supervised Learning} RAD-DINO~\citep{perez2024rad}, Medical MAE~\citep{xiao2023delving}, and C2L~\citep{zhou2020comparing} all leverage a self-supervised learning (SSL) objective without report supervision, which learns an image representation invariant to random augmentations. RAD-DINO and Medical MAE employ techniques such as masked image modeling~\citep{oquab2023dinov2}, while C2L uses a more classical contrastive learning approach.

\paragraph{Image-Text Alignment} BioViL-T \citep{bannur_learning_2023}, GLoRIA \cite{huang_gloria_2021} and BiomedCLIP~\citep{zhang2023biomedclip} use text supervision in a CLIP-style setup \citep{radford2021learning} by aligning image and text representation. In contrast to GLoRIA and BiomedCLIP,  which use only a single image and its associated text as input, BioViL-T additionally includes previous radiographs to account for potential backward references in the report. We also note that the BiomedCLIP model has been trained on more general biomedical images; hence, radiographs and medical reports constitute a minor portion of its training set.

\paragraph{Masked Image \& Text Modelling} Lastly, we include MRM~\citep{zhou2023advancing}, which is pre-trained with reports, but without explicit multi-modal alignment. Instead, given a subset of patches from an image, the model is trained to predict masked words in the report, as well as the remaining image patches. As such, its strategy combines image SSL and text supervision, encouraging the image representation to capture the contents of the report, while not being limited to it.

For the text encoder, we use the BERT model trained alongside BioViL-T in \citet{bannur_learning_2023} as it has been fine-tuned to better capture disease progression. In Appendix \ref{sec:ablations}, we evaluate the impact of the choice of text model.

\begin{table*}[h]
  \caption{An overview of the pre-trained models compared in this study. $\blacklozenge$ indicates that the image and text representations have been explicitly aligned. Size refers to the input image resolution.}
  \label{tab:backbone}
  \centering
  \begin{tabular}{llcccccc}
    \toprule
    & & & \multicolumn{2}{c}{Pre-Trained Using} & \multicolumn{3}{c}{Pre-Trained On}\\
    \cmidrule(r){4-5}     \cmidrule(r){6-8}
    Model ($V(x)$) & Type & Size & \small{Image SSL} & \small{Text} & \small{MIMIC} & \small{CheXpert} & \small{Other}\\
    \midrule
    RAD-DINO & ViT-B/14 & $518$ & \cmark  & \xmark & \cmark & \cmark & \cmark   \\
    C2L & ResNet-18 & $224$ & \cmark & \xmark & \cmark & \cmark & \cmark \\
    Medical MAE & ViT-B/16 & $224$ & \cmark & \xmark & \cmark & \cmark & \cmark \\
    MRM  & ViT-B/16 & $224$ & \cmark & \cmark & \cmark & \xmark & \xmark      \\
    BioViL-T  & ResNet-50 & $512$ & \xmark & \cmark$^\blacklozenge$ & \cmark & \xmark & \xmark \\
    GLoRIA  & ResNet-50 & $224$ &\xmark & \cmark$^\blacklozenge$ & \xmark & \cmark & \xmark \\
    BiomedCLIP 
    & ViT-B/16 & $224$ & \xmark & \cmark$^{\blacklozenge}$ & \xmark & \xmark & \cmark \\
    \bottomrule
  \end{tabular}
\end{table*}

\subsection{Fine-tuning}
\label{sec:3.2}

The goal during fine-tuning is to learn a classifier $f$ that requires only image representations $V(\mathbf{x})$ as the input and outputs the softmax-normalized probabilities in $\mathbb{R}^C$ for $C$-way classification. When fine-tuning \emph{without reports}, this is simply achieved by standard supervised learning, optimizing the parameters of $f$ to minimize the cross-entropy loss $\mathcal{L}$ computed over each $\bigl(f(\mathbf{x}_i), y_i \bigl)$ in the training dataset.
When fine-tuning \emph{with reports}, we adopt generalized distillation for privileged information (reports), as popularized by~\cite{lopez-paz_unifying_2016}---a two stage approach, described below.

In the first stage, we aim to learn a teacher model $g$ that takes $V(\mathbf{x})$ and $T(z)$ as the input, and outputs the probabilities, analogous to $f$. The objective is given by:
\begin{equation}
    \mathop{\mathrm{min}}_{g} \sum_{i=1}^N \left[ \mathcal{L} \Bigl(\, g \bigl(\,V(\mathbf{x}_i), T(z_i)\, \bigl) , y_i \Bigl) \right],
    \label{eq:1}
\end{equation}
where $N$ is the size of the training dataset. In the second stage, we train $f$ as a student model by distilling from the teacher $g$ according to the learning objective
\begin{align}
        \mathop{\mathrm{min}}_{f} & \sum_{i=1}^N \Big[ (1-\lambda)\mathcal{L} \Bigl(\, f \bigl(\,V(\mathbf{x}_i) \bigl) , y_i \Bigl) + \lambda \mathcal{L} \Bigl(\, f \bigl( V(\mathbf{x}_i) \bigl), g^\tau \bigl(\,V(\mathbf{x}_i), T(z_i)\bigl) \Bigl) \Big],
    \label{eq:2}
\end{align}
where $\lambda\in[0,1]$ is the imitation parameter that balances the importance of following the teacher prediction, and $g^\tau(\cdot)$ denotes the teacher prediction computed with a softmax temperature of $\tau > 0$. 

Previous works have observed that results are rarely sensitive to $\lambda$ \citep{yang2022toward}, and we found $\lambda = 0.75$ to work well in our experiments. In Appendix \ref{sec:ablations} we include the ablation of $\tau$.
Overall, this setup trains the image-only classifier $f$ to imitate a privileged teacher $g$ that has access to reports, thereby guiding $f$ to classify $\mathbf{x}$ like a radiologist without accessing $z$ at test time.
For direct comparison, we also implement a self-distillation baseline, which is trained without reports by distilling from an image-only teacher model.

\noindent\textbf{Fine-tuning architecture.} We design $f$ to first perform self-attention on the visual tokens output from $V$, where the query, key and value networks are learnable linear heads. Then we perform mean pooling on the self-attention tokens output and feed the resulting feature into a learnable linear classifier. The teacher network $g$ processes the text tokens output from $T$ with a separate self-attention and mean pooling step. Then, the pooled visual and text features are concatenated as the input to a single linear classifier.
We choose this design as it performs well empirically (comparisons in Appendix \ref{sec:ablations}), without over-complicating the fine-tuning stage.
\section{Experimental Setup}
\label{sec:4}

Next, we detail and motivate the design choices behind our experiments aimed at aiding understanding of the benefits of using radiology reports in pre-training and fine-tuning of image-only classifiers. 

\subsection{The diversity of image classification tasks}
\label{sec:categories}

\begin{figure*}
    \centering
    \footnotesize
    \begin{subfigure}[t]{\textwidth}
         \includegraphics[width=\textwidth]{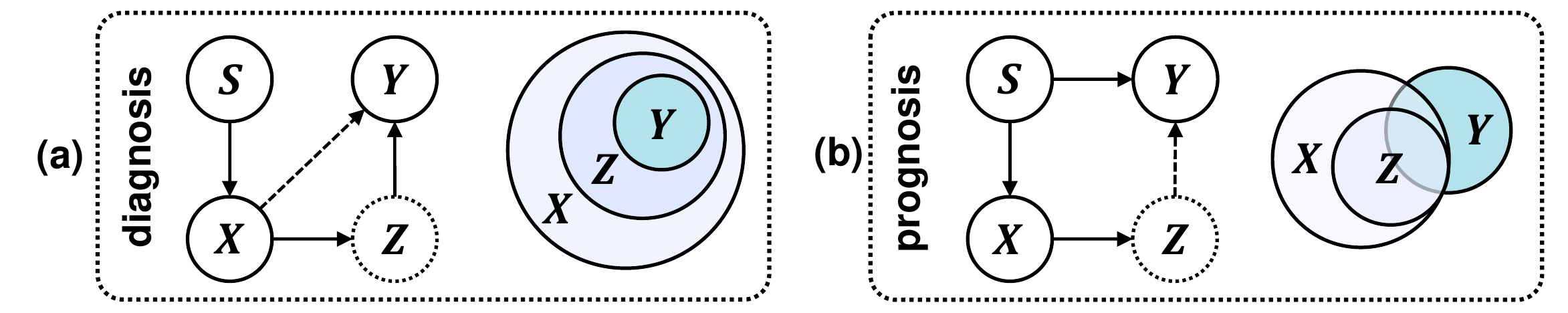}
         \phantomcaption
         \label{fig:3a}
    \end{subfigure}
    \begin{subfigure}[t]{0\textwidth} 
         \includegraphics[width=0\textwidth]{example-image-b}
         \phantomcaption
         \label{fig:3b}
    \end{subfigure}
    \vspace*{-9mm}
    \caption{Example causal graphs under the (a) diagnosis or (b) prognosis setting, where $S$: (unobserved) patient state, $X$: medical image, $Z$: radiology report (potentially missing at test time) and $Y$: target label. Dashed edges indicate associations likely to be weak. The accompanying Venn diagrams conceptually illustrate the relationship between the information contained in the observed variables.}
    \vspace*{-4mm}
    \label{fig:causal_graph}
\end{figure*}

The vast majority of work in medical image classification has focused on \emph{diagnostic tasks}, where the label $Y$ represents the presence of one or more medical conditions~\citep{johnson2019mimic, bustos2020padchest, irvin2019chexpert}. This ignores that clinical prediction challenges frequently involve \emph{prognostic} targets such as future (\eg, 30-day) mortality \citep{nam2022deep}, hospital readmission \citep{huang2019clinicalbert}, and disease forecasting \citep{li_causal_2021}. Critically, the causality and association strength between radiograph $X$, report $Z$, and target variable $Y$ can vary substantially depending on the nature of the task. We illustrate this in Figure~\ref{fig:causal_graph} and give examples below. 
In many \emph{diagnostic} tasks, like pneumonia detection~\citep{rajpurkar2017chexnet}, the target label $Y$ is determined almost completely by the radiology report $Z$ (see Figure~\ref{fig:3a}).  Often, the report will contain mentions of a likely diagnosis, and if not, it is written to aid physicians in making one. 
This is taken to its extreme in currently widely-used benchmarks where the label $Y$ has been extracted from $Z$ based on a set of hand-written rules \citep{bustos2020padchest, johnson2019mimic, irvin2019chexpert}. In other words, $Y$ is deterministically caused by $Z$.
In contrast, \emph{prognostic} tasks, such as predicting 3-day discharge or 12-month readmission~\citep{kansagara2011risk}, the target variable $Y$ is often highly stochastic relative to both the image $X$ and report $Z$ (Figure~\ref{fig:3b}). Outcomes depend largely on latent disease progression and external factors, while radiographs and reports play only a limited, indirect role by informing treatment decisions.

Figure~\ref{fig:causal_graph} does not represent all categories of medical image classification. Auxiliary tasks like predicting the age of a patient~\citep{perez2024rad} 
fit in neither the prognostic nor diagnostic category (we label them \emph{auxiliary}), and their causal graphs are harder to determine. 
Yet, it is clear that the nature of the task affects how much information the report $Z$ carries about $Y$, and whether it can be extracted from $X$. Consequently, \emph{we need richer benchmarks to fully understand the effectiveness of utilizing reports in pre-training and fine-tuning of medical image classifiers}. 

\subsection{Benchmark and evaluation details}

We compile our experiments from two existing datasets, MIMIC-CXR \citep{johnson2019mimic} and INSPECT \citep{huang2023inspect}, as they both contain radiology reports. INSPECT is a multimodal dataset containing CT scans of patients at risk of pulmonary embolism. We reconstruct chest X-rays from these CT volumes using digital radiograph reconstruction\footnote{The code for this will be made available}. None of the image encoders have been trained on this dataset, so it offers a fair test of generalization. We refer to Appendix \ref{sec:data} for more information regarding this, preprocessing of images and reports, and dataset construction.

1) MIMIC-CXR, where we consider the following tasks:
\begin{itemize}[itemsep=1pt, left=0pt]
    \item {MIMIC 5x1200 - \emph{Diagnostic} (6,000 images):} We perform the common task of predicting diagnostic labels extracted from radiological reports. To do this, we construct MIMIC CXR 5x1200 (similar to CheXpert 5x200 in \citet{huang_gloria_2021}) by choosing a subset of the MIMIC-CXR-JPG subjects that have had exactly one of five labels assigned to them. Following \citet{irvin2019chexpert}, these labels are Atelectasis, Cardiomegaly, Edema, Pleural Effusion, and Consolidation. The final dataset includes $1200$ images per label ($1000$ for training and $200$ for evaluation).

    \item{Age - \emph{Auxiliary} (35,242 images):} To evaluate model performance on targets rarely discussed in the report, we predict the age of a patient based on their radiograph. We construct a classification problem by dividing the ages into 5 bins, as done in \citet{perez2024rad}. For each seed, $90\%$ of images are sampled for training, with $10\%$ withheld for validation. 

    \item{3-Day Discharge - \emph{Prognostic} (18,490 images):} By linking the images in MIMIC-CXR to the patient records in MIMIC-IV \citep{johnson2023mimic}, we gain access to admission information for each patient. As a short-term prognostic target, we predict whether a patient will be discharged in the coming 3 days. For each seed, $90\%$ of images are used for training, and $10\%$ withheld for validation.
\end{itemize}

2) INSPECT, where we consider the following tasks:

\begin{itemize}[itemsep=1pt, left=0pt]
    \item 12-Month PH - \emph{Prognostic} (5,449 images): A binary classification task where $1$ indicates that a patient was diagnosed with pulmonary hypertension (PH) within 12 months of an image being taken. The dataset is collected by removing the censored samples and then artificially balancing it by subsampling the number of patients that did not experience the event. For each seed, $75\%$ of images are sampled for training, while $25\%$ are withheld for validation.  
    \item Readmission - \emph{Prognostic} (5,651): A binary classification task, predicting whether a patient will be readmitted in the coming 12 months. Censored patients were removed, and the dataset was artificially balanced similar to 12-Month PH. For each seed, $75\%$ of images are sampled for training, while $25\%$ are withheld for validation. 
\end{itemize}

We fine-tune the 6 models covered in Section \ref{sec:3} on each of these datasets. All models are trained with a learning rate of $10^{-4}$ using the Adam optimizer and a batch size of 64. Performance is evaluated by measuring AUC after every epoch, with one-versus-rest and micro averaging for the multiclass problems. When training the teacher model, we save the version with the highest AUC on the validation set. We used the distillation temperature parameter $\tau = 0.25$ for all datasets except MIMIC 5x1200, which used $\tau = 2.5$. Experiments are averaged over multiple seeds with different network parameter initializations and training subsets. More detailed  information in Appendix \ref{sec:data}.

\section{Related Work}
\label{sec:2}

\textbf{Evaluation of medical image models} is challenging due to differences in preprocessing, training, and validation datasets. \citet{zhou2024benchx} introduce BenchX to standardize evaluation of medical Vision Language Models (VLMs), but focus on pre-training representations and tasks like segmentation and report generation rather than comparing with self-supervised methods. \citet{khader2023medical} show ICU survival prediction improves when combining chest X-rays with electronic health records, while \citet{weissman2018inclusion} demonstrate gains from including clinical notes. \citet{castro2020causality} emphasize the importance of understanding the causal relationship between medical images and their annotations and argue that there is often a mismatch between the training dataset and the target application.

\textbf{Learning using privileged information} allows models access to additional features during training but not at test time \citep{vapnik2009new}. Generalized distillation was popularized by \citet{lopez-paz_unifying_2016} and has been applied to recommendation systems using richer supervision \citep{yang2022toward, xu2020privileged}. Transfer and Marginalize (TRAM) \citep{collier2022transfer} addresses noisy labels, with \citet{ortiz2023does} showing PI helps models avoid overfitting shortcuts. Other work uses intermediate steps in time-series as PI \citep{jung2022efficient, karlsson2022using}, improving sample efficiency by learning transformations between them.

\section{Results}
\label{sec:5}

\begin{figure*}[t]
\centering
     \resizebox{0.9\textwidth}{!}{
        \begin{minipage}{\textwidth}

    \begin{subfigure}[T]{0.33\textwidth}
        \centering
        \includegraphics[width=0.99\linewidth]{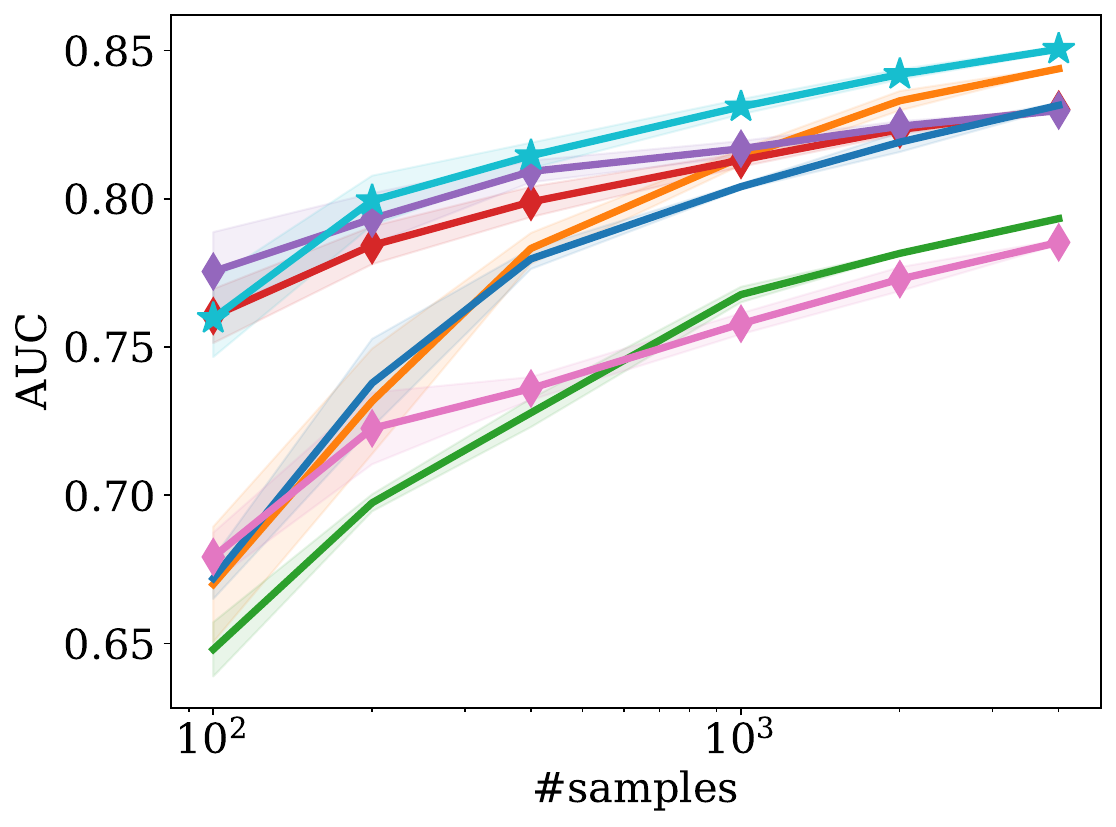}
        \caption{\textbf{MIMIC 5x1200 - Diagnostic}}
        \label{fig:report_label}
    \end{subfigure}
    \hfill
    \begin{subfigure}[T]{0.33\textwidth}
       \centering
        \includegraphics[width=0.99\linewidth]{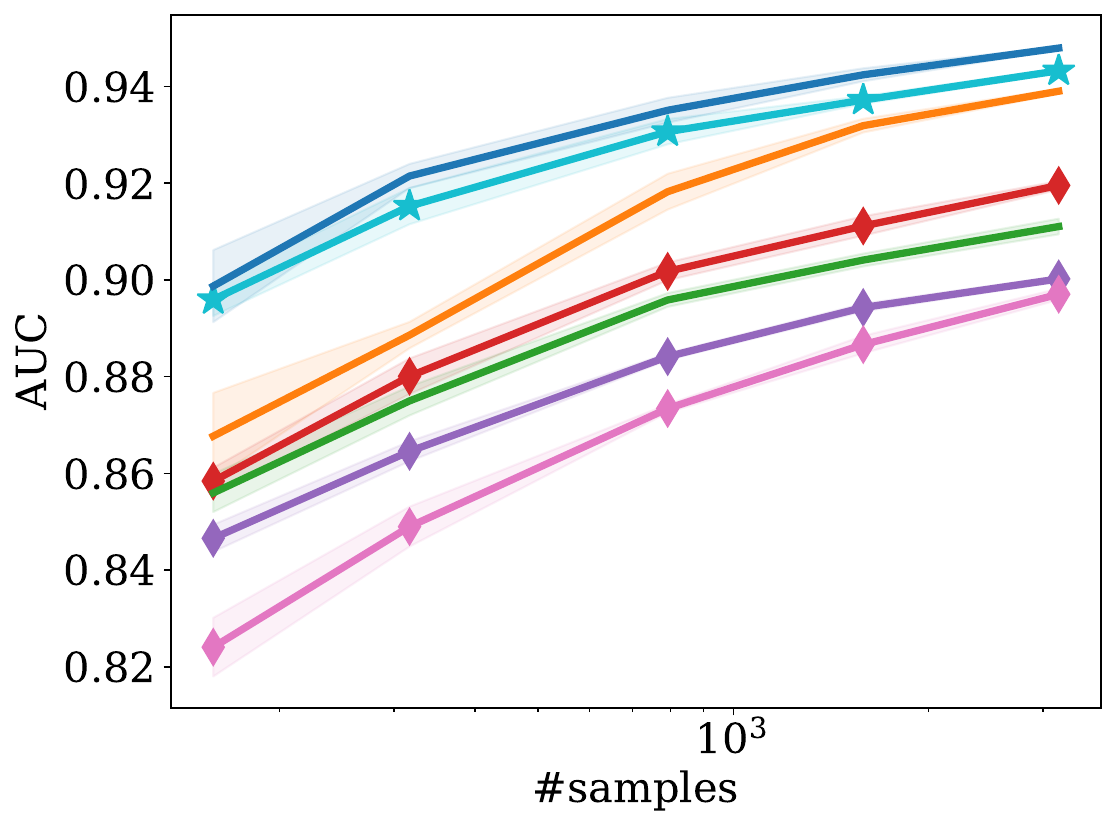}
        \caption{\textbf{Age - Auxiliary}}
        \label{fig:age}
    \end{subfigure}
    \centering
    \begin{subfigure}[T]{0.33\textwidth}
        \centering
        \includegraphics[width=0.99\linewidth]{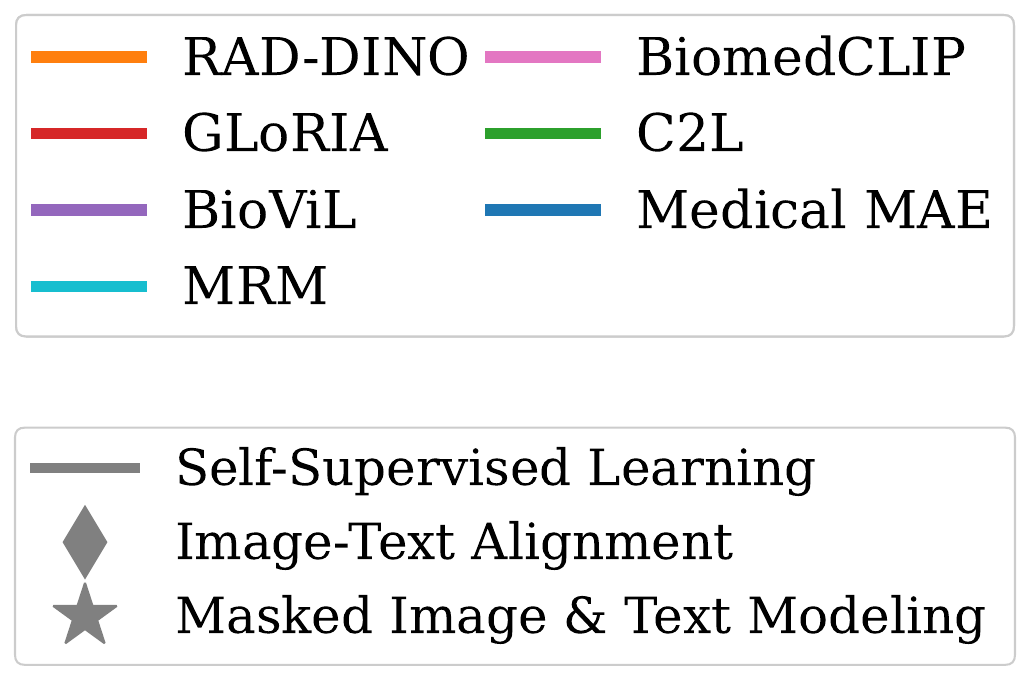}
    \end{subfigure}
    \hfill
    \begin{subfigure}[T]{0.33\textwidth}
        \centering
         \includegraphics[width=0.99\linewidth]{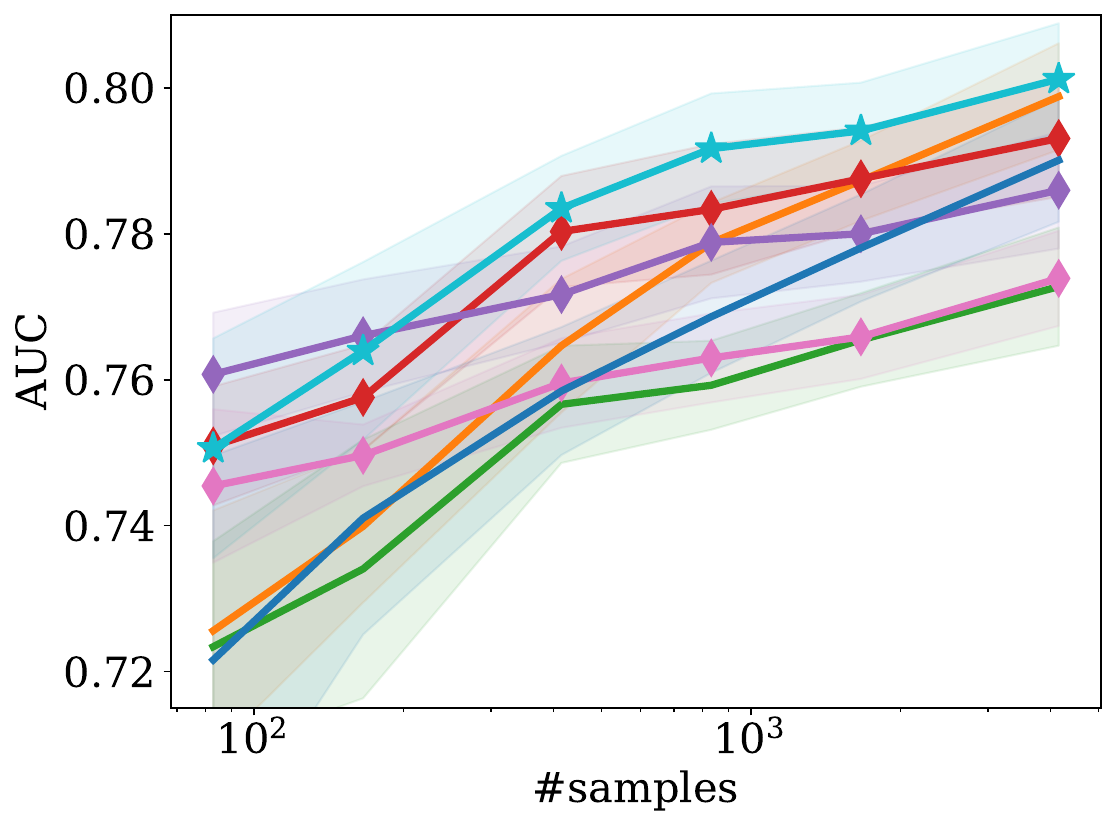}
        \caption{\textbf{3-Day Discharge - Prognostic}}
        \label{fig:dischage_backbones}
    \end{subfigure}
    \hfill
    \begin{subfigure}[T]{0.33\textwidth}
       \centering
        \includegraphics[width=0.99\linewidth]{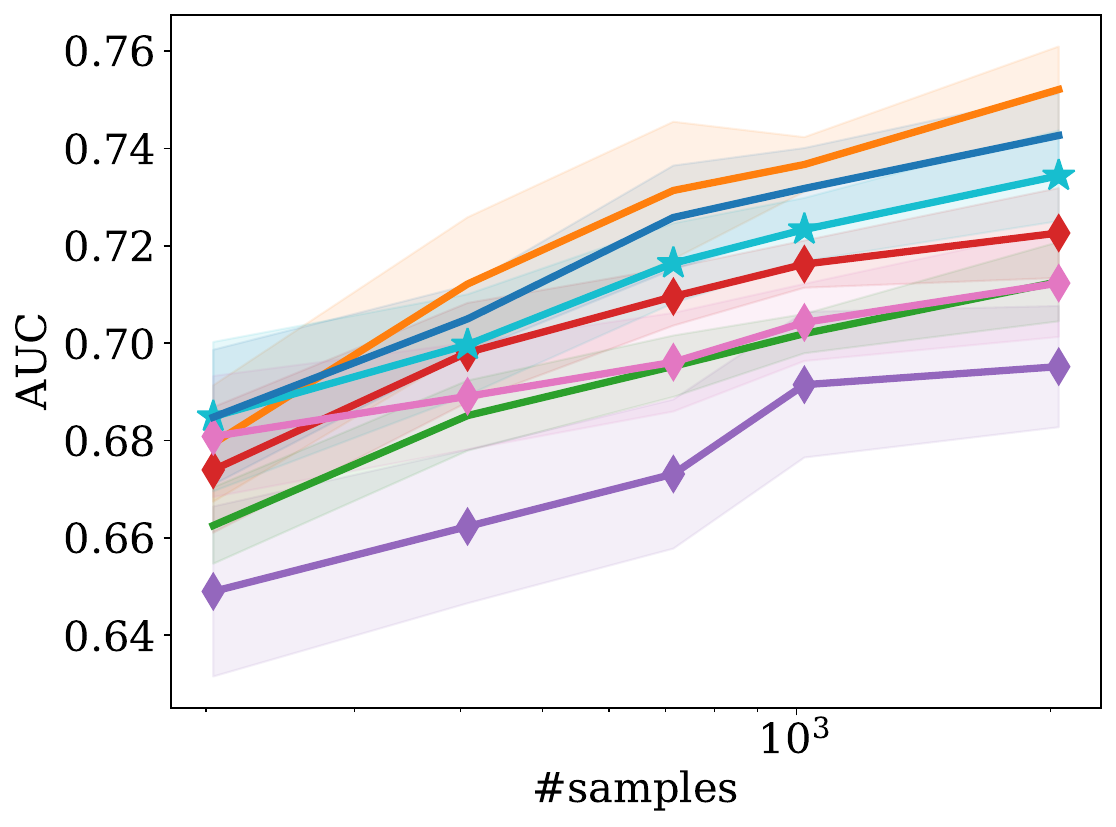}
        \caption{\textbf{12-Month PH - Prognostic}}
        \label{fig:12_month}
    \end{subfigure}\hfill
    \begin{subfigure}[T]{0.33\textwidth}
       \centering
        \includegraphics[width=0.99\linewidth]{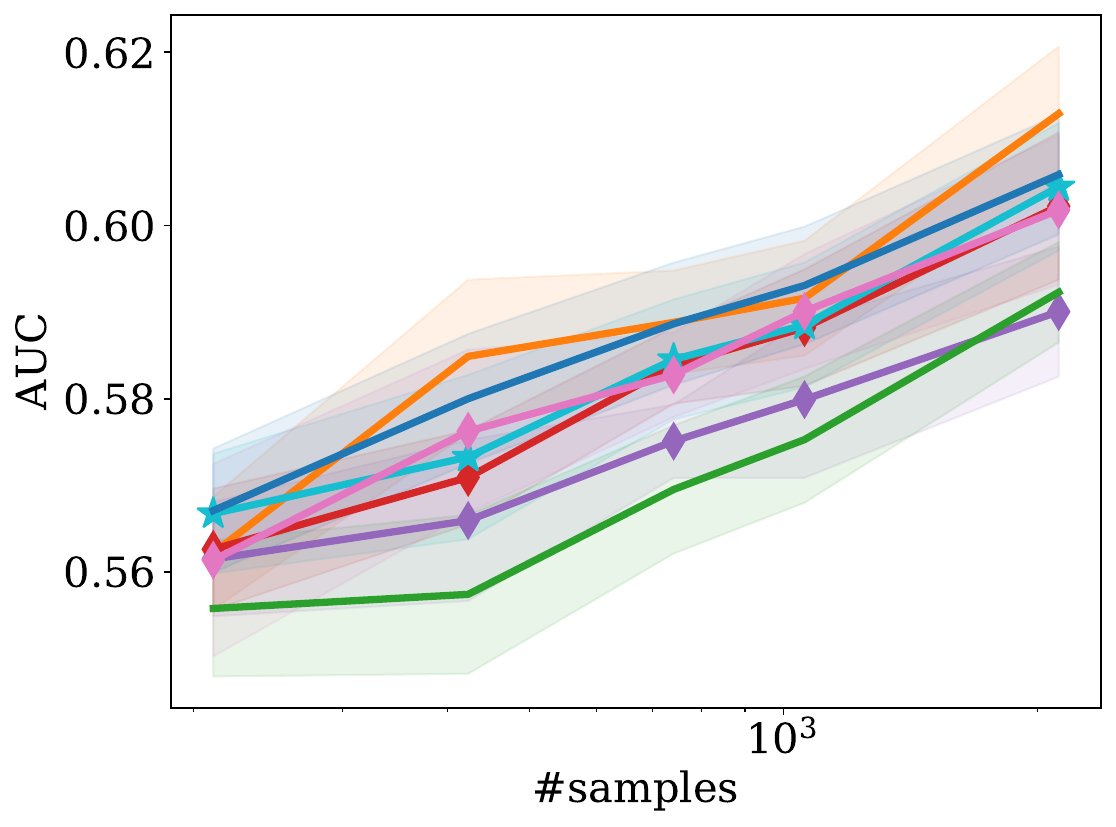}
        \caption{\textbf{Readmission - Prognostic}}
        \label{fig:readmin}
    \end{subfigure}
    \end{minipage}
    }
    \caption{A comparison of the sample efficiency of different backbones. The plots are averaged over 5 seeds, except for Readmission, which is averaged over 10. The shaded area regions represent the 90\% CI. The x-axis is on a logarithmic scale. AUPRC results are included in Appendix \ref{sec:ext_results}}
    \label{fig:backbone_comp}
\end{figure*}
    
\begin{table*}[h]
  \caption{A comparison of text-only and image-only models. Text results come from fine-tuning the BioViL-T BERT model, ''Image AUC'' corresponds to the best-performing model.}
  \label{tab:text}
  \centering
  \begin{tabular}{lccccc}
    \toprule
    Dataset & \textbf{M 5x1200} & \textbf{Age} & \textbf{3-Day} & \textbf{12-Month PH} & \textbf{Readmission} \\
    \midrule
    Fraction & $5\%$ & $1\%$ & $1\%$ & $5\%$ & $5\%$ \\
    \midrule
    Text AUC (sd) & $\mathbf{96.4}$ ($0.0$) & $78.5$ ($0.9$) & $72.5$ $(1.5)$ & $\mathbf{71.0}$ ($1.7$) & $\mathbf{57.4}$ ($1.9$)\\
    Image AUC (sd) & $79.9$ ($1.1$) & $\mathbf{92.1}$ ($0.3$) & $\mathbf{77.4}$ ($0.9$) & $68.5$ ($2.1$) & $56.7$ ($1.4$)\\
    Image Model & MRM & Medical MAE & BioViL & MRM & Medical MAE\\
    \midrule
    Fraction & $50\%$ & $10\%$ & $10\%$ & $50\%$ & $50\%$ \\
    \midrule
    Text AUC (sd) & $\mathbf{98.5}$ ($0.0$) & $80.6$ ($0.2$) & $76.0$ ($0.8$) & $\mathbf{76.6}$ ($0.8$)& $60.7$ ($1.4$) \\
    Image AUC (sd) & $84.2$ ($0.2$) & $\mathbf{94.8}$ ($0.0$) & $\mathbf{79.4}$ ($0.9$) & $75.2$ ($1.2$) & $\mathbf{61.3}$ ($1.5$)\\
    Image Model & MRM & Medical MAE & MRM & RAD-DINO & RAD-DINO\\
    \bottomrule
  \end{tabular}
\end{table*}

\subsection{Impact of pre-training with reports}

Recently, text-based supervision during pre-training has been criticized \citep{perez2024rad} as explicit alignment has been shown to decrease the generality of visual representations \citep{zhai2022lit, yang2022vision}. In Figure \ref{fig:backbone_comp}, we perform experiments to see if and when this is noticeable in the medical image classification setting. To gauge the predictiveness of the reports (i.e., how well we can predict $Y$ from $Z$), we include Table \ref{tab:text}, which compares a text-only model against the best-performing image-only model.

\begin{figure*}[ht]
    \centering
     \resizebox{0.85\textwidth}{!}{
        \begin{minipage}{\textwidth}

    \centering
    \includegraphics[width=0.99\textwidth]{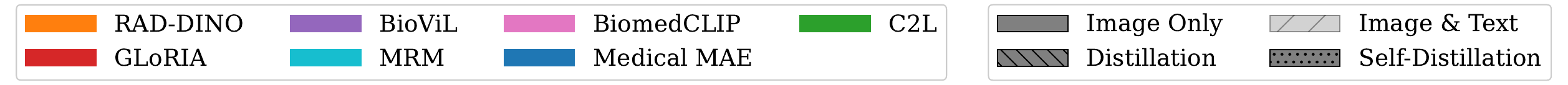}
    
    \centering
    \begin{subfigure}[B]{0.64\textwidth}
        \centering
        \includegraphics[width=0.54\textwidth]{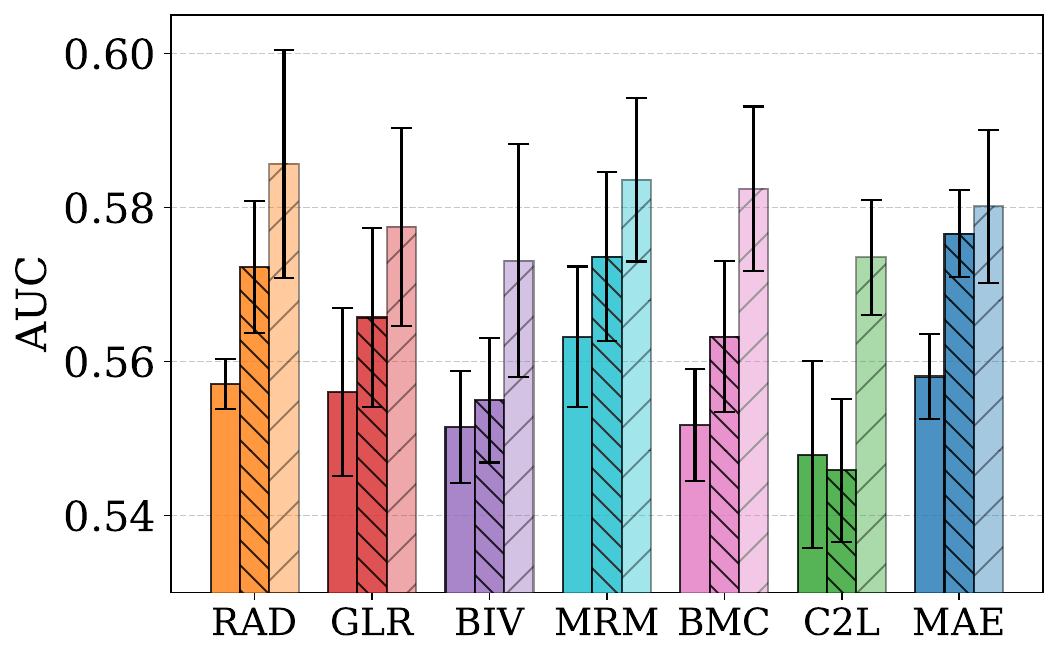}
        \includegraphics[width=0.45\textwidth]{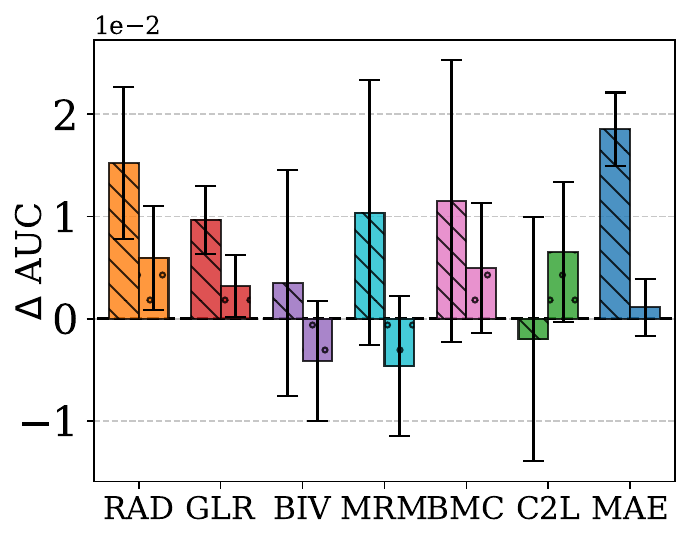}
        \caption{$5 \%$ of \textbf{Readmission} (211 samples) Left: Average AUC \\Right: Average AUC-difference  between distillation and image-only}
        \label{fig:readm_0.05}
    \end{subfigure}\hfill
    \hfill
    \begin{subfigure}[B]{0.345\textwidth}
        \centering\includegraphics[width=\textwidth]{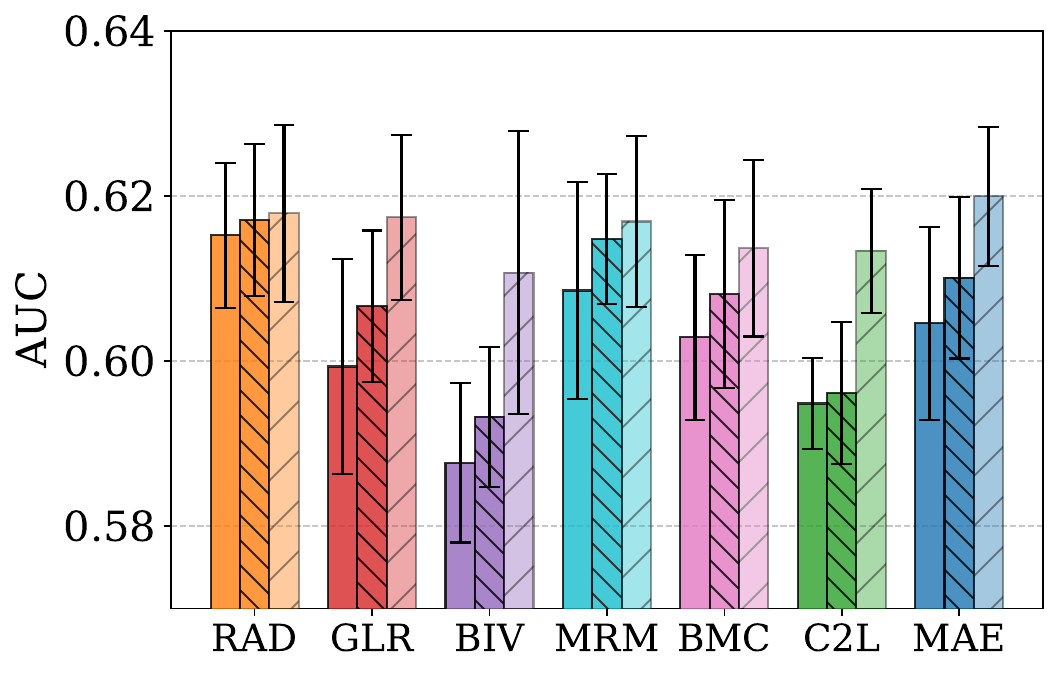}
        \caption{$50 \%$ of \textbf{Readmission} (2119 samples) Average AUC}
        \label{fig:readm_0.5}
    \end{subfigure}
    \centering
    \centering
    \begin{subfigure}[B]{0.64\textwidth}
        \centering
        \includegraphics[width=0.54\textwidth]{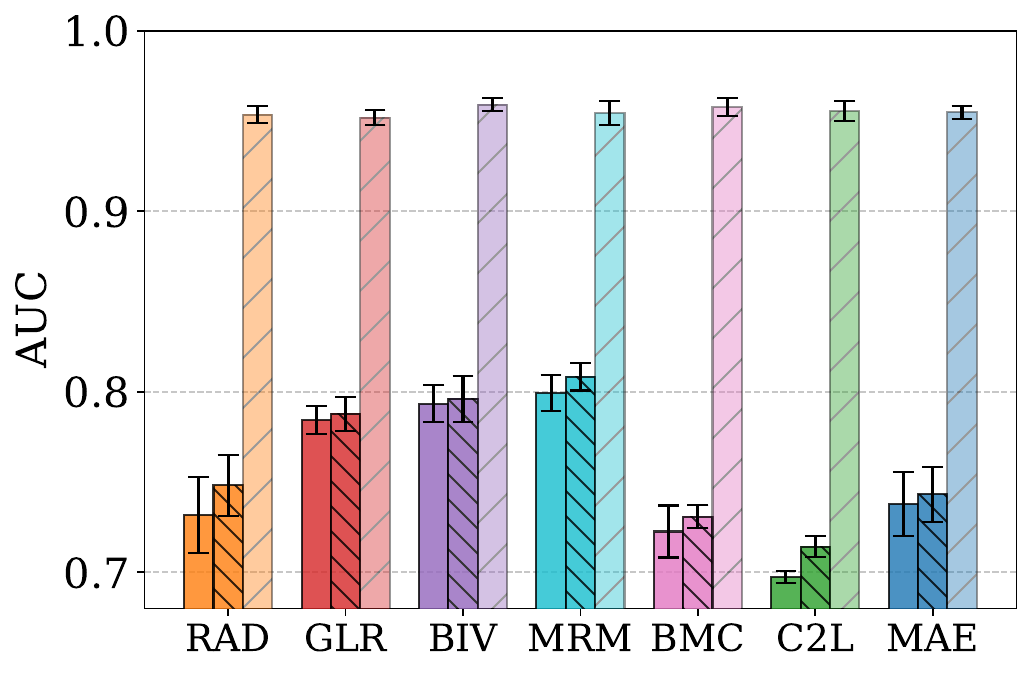}
        \includegraphics[width=0.45\textwidth]{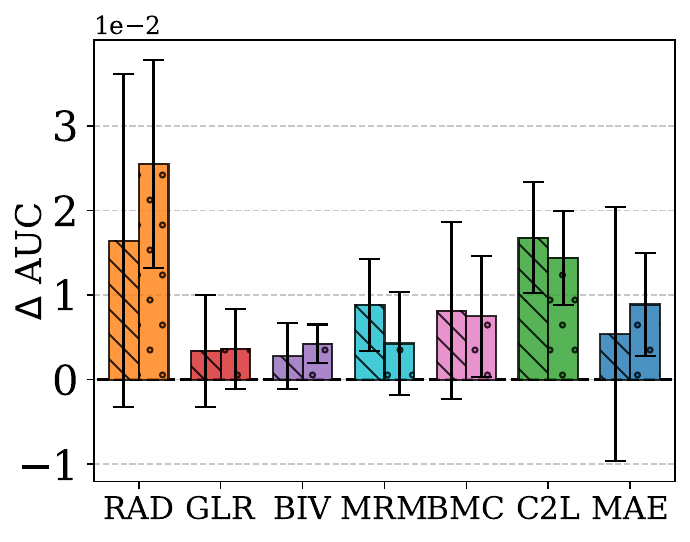}
        \caption{$5 \%$ of \textbf{MIMIC 5x1200} (200 samples) Left: Average AUC \\Right: Average AUC-difference  between distillation and image-only}
        \label{fig:report_labels_self_dist}
    \end{subfigure}\hfill
    \hfill
    \begin{subfigure}[B]{0.345\textwidth}
        \centering\includegraphics[width=\textwidth]{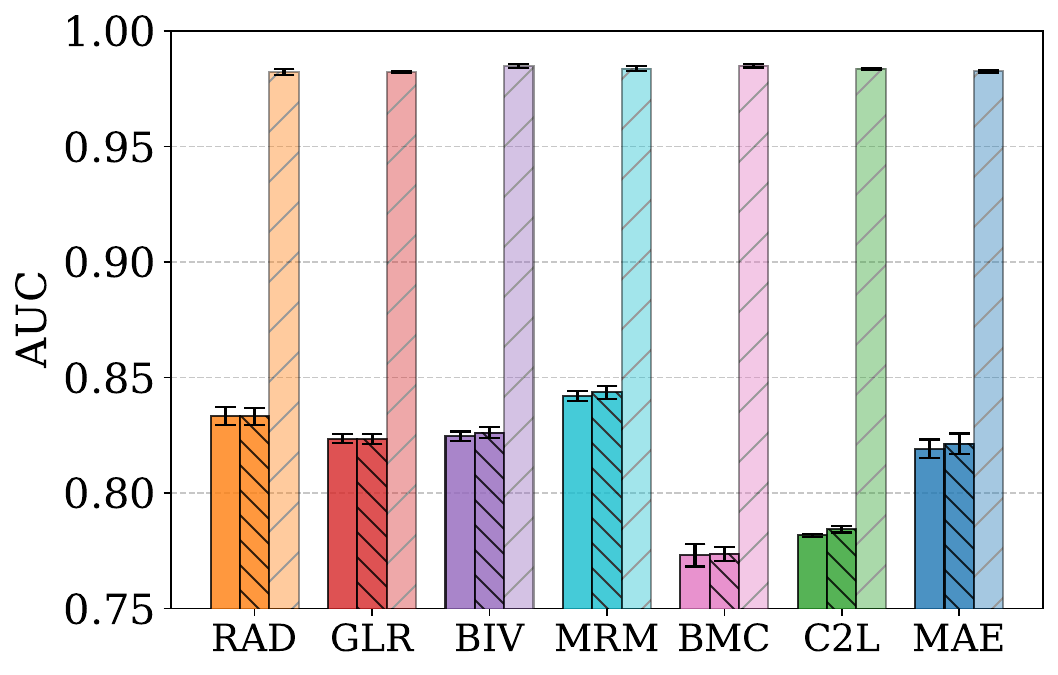}
        \caption{$50 \%$ of \textbf{MIMIC 5x1200} (2000 samples) Average AUC}
        \label{fig:report_0.5}
    \end{subfigure}
    \end{minipage}
    }
    \caption{Distillation results on \textbf{Readmission} (prognostic) and \textbf{MIMIC5x1200} (diagnostic) with different training sizes, averaged over 5 seeds. Bars represent the 95\% CI. AUPRC results are found in Appendix \ref{sec:ext_results}.}
    \label{fig:barplot}
\end{figure*}

\paragraph{Explicitly aligning with reports limits generalizability.} As expected (see Section~\ref{sec:categories}), we observe the highest relative performance of GLoRIA and BioViL compared to other models in the \textbf{MIMIC 5x1200} experiment (Figure \ref{fig:report_label}), where the labels have been extracted from the reports.  This is mainly evident in the low sample regime, and RAD-DINO overtakes both when the number of samples increases to over $1,000$. However, when predicting \textbf{Age} (still using images from the MIMIC dataset), the relative performance of these algorithms decreases significantly compared to backbones pre-trained with self-supervision. Instead, Medical MAE, MRM, and RAD-DINO perform the best, with the previously poorly performing C2L achieving a higher AUC than BioViL, and only slightly lower than GLoRIA. The \textbf{Age} experiment has the largest gap between image-only and text-only performance in Table \ref{tab:text}, which offers a possible explanation as to why aligning with reports is not beneficial for this task. Finally, BiomedCLIP, which has been trained on more general biomedical images and their captions, performs poorly in both settings.

\paragraph{Image self-supervision is usually beneficial for prognostic tasks.} For the prognostic tasks, the order of performance varies. In \textbf{3-Day Discharge} (Figure \ref{fig:dischage_backbones}), GLoRIA, BioViL, and MRM (all pre-trained with text supervision) perform well in the low sample regime. Yet, in the \textbf{12-Month PH} experiment (Figure \ref{fig:12_month}), GLoRIA and BioViL do not see the same benefits, with BioViL performing the worst in all sample sizes. The trend is similar in \textbf{Readmission} (Figure \ref{fig:readmin}). We hypothesize that while these tasks are prognostic, whether or not a patient is discharged in the coming days may be more closely correlated to a diagnostic label discussed in the radiology reports. For example, it is well possible for pneumothorax (one of the labels extracted from reports in MIMIC-CXR \citep{johnson2019mimic}) to persist for more than 3 days \citep{thachuthara2021pneumothorax}. Lastly, the high sample efficiency of MRM in all experiments highlights the benefit of not explicitly aligning representations with text, but using text supervision as a complement to image self-supervision.

\subsection{Fine-tuning with reports}
\begin{figure*}[h]
\centering
    \begin{subfigure}[T]{0.4\textwidth}
       \centering
        \includegraphics[width=0.95\linewidth]{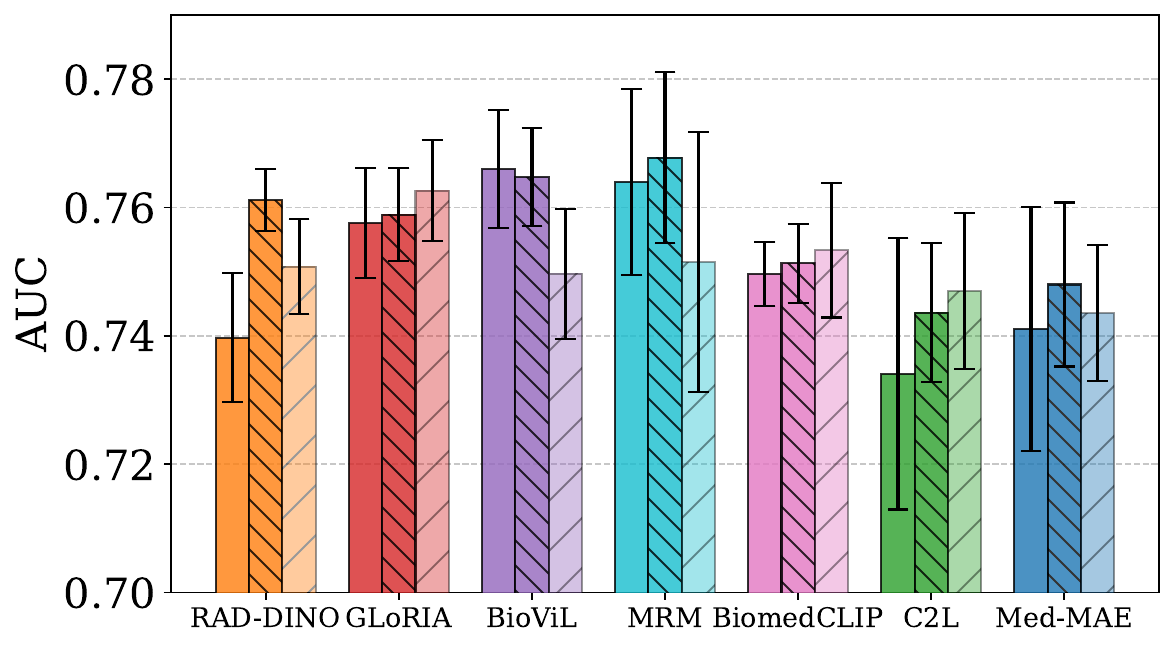}
        \caption{$1 \%$ of training data \\(165 samples)}
        \label{fig:discharge_box}
    \end{subfigure}
    \begin{subfigure}[T]{0.38\textwidth}
        \centering
        \includegraphics[width=0.95\linewidth]{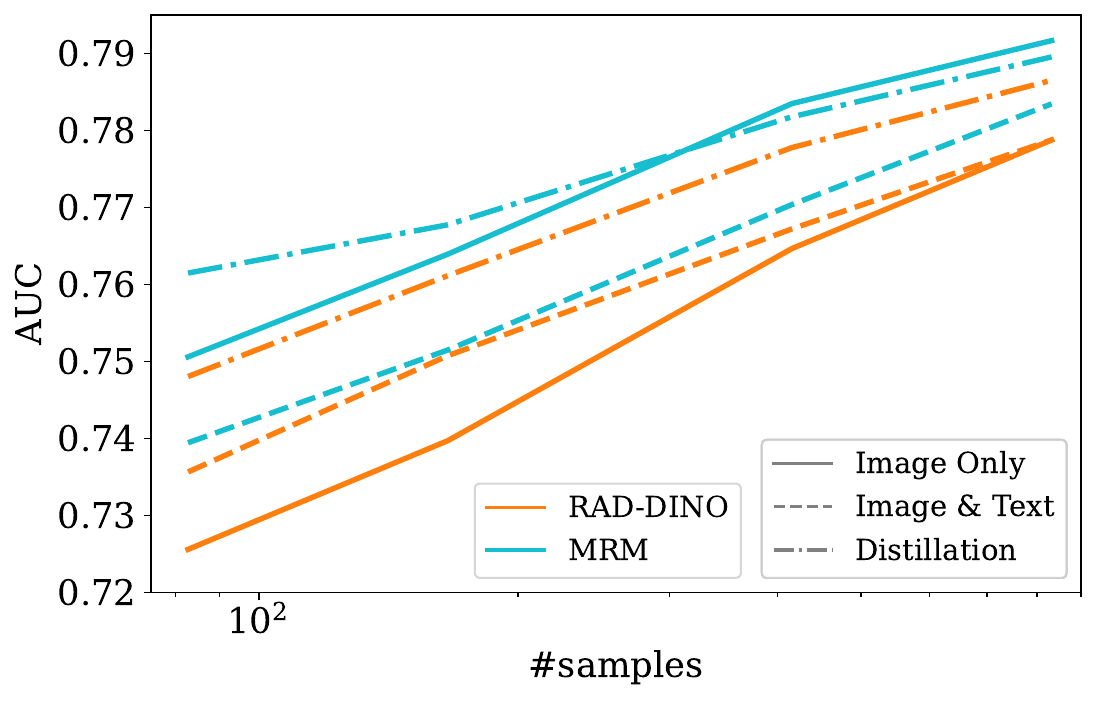}
        \caption{As a function of training set size}
        \label{fig:discharge_dino}
    \end{subfigure}
    \caption{Mean AUC, averaged over 5 seeds, when performing distillation on \textbf{3-Day Discharge}.}
    \label{fig:outperform}
\end{figure*}
    
Leveraging radiology reports during pre-training as a means to achieve better data efficiency has been widely studied \citep{zhang_contrastive_2022, huang_gloria_2021, bannur_learning_2023}. At the same time, the use of these texts during fine-tuning has been ignored, leaving their possible usefulness for this unknown. We perform experiments on \textbf{MIMIC 5x1200}, \textbf{3-Day Discharge}, and \textbf{Readmission} (Figures~\ref{fig:barplot} and ~\ref{fig:outperform}) to explore whether distilling from a teacher with access to text can have a similar impact. Due to space constraints, results for \textbf{Age} and \textbf{12-Month PH} are provided in Appendix \ref{sec:ext_results}.

\paragraph{Distillation can have a larger impact than pre-training.} Interestingly, pre-training does not always have the largest effect on performance. In Figure \ref{fig:readm_0.05}, distillation has a higher impact on the AUC than the choice of pre-training method when predicting \textbf{12-Month Readmission} in the small-sample domain. The results show that distillation from a teacher with access to the text report consistently increases the AUC across seeds, something that is not observed when applying self-distillation from an image-only model. As the number of samples grows (Figure \ref{fig:readm_0.5}), distillation still leads to a meaningful performance increase, but the gap between different image backbones widens.

\paragraph{Distillation works poorly if the text is too predictive.} On the other hand, the results of distillation on the \textbf{MIMIC 5x1200} dataset (Figures~\ref{fig:barplot}(c--d)) highlight that a strong multi-modal teacher does not imply that the student will benefit from distillation. The student sees no more benefit in distilling from the significantly more accurate teacher as opposed to performing self-distillation (Figure~\ref{fig:report_labels_self_dist}). This aligns with previous research, which has observed that utilizing PI through distillation performs poorly if the information is too predictive of the label \citep{yang2022toward, ortiz2023does}. The results indicate that this is the case for diagnostic labels of this kind, further underscoring that the benefit from using reports for pre-training as opposed to generalized distillation depends on the task structure (as hypothesized in  Section~\ref{sec:categories}). The high performance of models pre-trained with text in Figure~\ref{fig:report_label} brings into question whether an alternate fine-tuning method might better leverage reports in this setting, allowing self-supervised models to catch up.

\paragraph{Benefits of distillation are dependent on both task and backbone.} We don't always benefit from applying distillation, even in the prognostic setting. Beyond variance with the task, our results show that not all encoders benefit the same when fine-tuning with PI. Medical MAE, RAD-DINO, MRM and GLoRIA see large performance increases in the \textbf{12-Month Readmission} experiment (Figure \ref{fig:readm_0.05}), while BioViL and C2L do not improve. Similarly, RAD-DINO benefits more than any other model in the \textbf{3-Day Discharge} setting (Figure \ref{fig:outperform}), and GLoRIA does not seem to benefit at all in contrast to before. The benefits are expected to depend on the relation between image ($X$), text ($Z$), and label ($Y$). In practice, we have both the image and text backbones frozen, meaning that the performance will depend on the relation between V(x) and T(z). 

\paragraph{An image-only student can outperform a privileged teacher.} Lastly, the multi-modal teacher can perform worse than the image-only baseline without access to the report (Figure \ref{fig:discharge_box}) in the low sample regime.
This is possibly a result of the increase in the number of features and model complexity (introduced by the addition of a new modality), which can decrease performance in low-sample settings~\citep{huang2021makes}.
Despite this, in the \textbf{3-Day Discharge} experiment, we observe that the student can outperform the teacher it is distilling from (Figure \ref{fig:discharge_dino}). While this may be surprising in the PI setting, a student network outperforming its teacher has been observed numerous times in literature ~\citep{furlanello2018born, pareek2024understanding, yang2022toward} and forms the foundation of self-distillation.
In Appendix \ref{sec:ext_results}, we show results for each seed, demonstrating that RAD-DINO consistently outperforms the multi-modal model with distillation.

\section{Conclusion}
\label{sec:conclusion}

We have studied the utility of radiology reports in the pre-training and fine-tuning of medical image-only classifiers. To address the narrow focus of current classification benchmarks on diagnostic targets, we compiled a collection of diagnostic, prognostic, and auxiliary tasks.
We compared the performance of 6 image encoders, pre-trained with or without radiology reports when fine-tuned for 5 image-only classification tasks, with or without reports during fine-tuning. 
Our results underscore the importance of task-diversity in evaluation to better understand the pros and cons of different methods. We find that 1) Pre-training with text shows the greatest benefit for diagnostic labels, but that 2) explicit text-alignment during pre-training can lose information about targets weakly associated with reports. Further, 3) fine-tuning with distillation from image-report models can have impact comparable to the choice of image encoder, but benefits depend on the encoder and task. 

Our study is limited to using text as PI during fine-tuning with a single text encoder and a single learning objective (distillation). While our goal is to highlight the potential of leveraging radiology reports during fine-tuning, rather than identifying the best possible PI method, the fact that distillation does not improve performance even when the text is completely predictive (MIMIC 5x1200) indicates that future work may benefit from developing more refined approaches. 


\section*{Acknowledgements}

We thank Ida Häggström for her advice regarding digital radiograph reconstruction and for suggesting the Plastimatch software suite. We would also like to thank Victor Wåhlstrand Skärström for his input and valuable discussion at the start of this project. FDJ and HB are supported by Swedish Research Council Grant 2022-04748. FDJ is also supported in part by the Wallenberg AI, Autonomous Systems and Software Program, founded by the Knut and Alice Wallenberg Foundation. YZ is funded by the Wallenberg-NTU Presidential Postdoctoral Fellowship. The computations were enabled by resources provided by the National Academic Infrastructure for Supercomputing in Sweden (NAISS), partially funded by the Swedish Research Council through grant agreement no. 2022-06725.
\bibliographystyle{plainnat}
\bibliography{main}
\clearpage
\appendix

\section*{Appendix}

This appendix includes additional experimental results and information about the training setup. Further details about data processing, backbone model usage, and hyperparameters are presented in Section \ref{sec:data}. Section \ref{sec:ext_results} contains the results from distillation experiments on \textbf{Age} and \textbf{12-Month PH}, and an additional diagnostic experiment on the INSPECT dataset that reaffirms the limited benefit of distillation when reports are too predictive. In Section \ref{sec:ablations} we perform multiple ablation experiments that motivate hyperparameter choices, while showcasing the robustness of our results. Sections \ref{sec:compute} and \ref{sec:license} cover compute usage and dataset licenses, respectively. 

\section{Training \& Data Processing}
\label{sec:data}

When processing radiology reports in MIMIC-CXR we extract the impressions section and, if available, the findings section. INSPECT is a multimodal dataset containing CT images and pre-extracted impression sections from their accompanying reports. We perform digital radiograph reconstruction using the Plastimatch software suite \citep{sharp2010plastimatch} to convert the CT volumes to (anterior-posterior) radiographs that our pre-trained backbones can process. After extracting the radiograph, we follow the preprocessing of \citet{johnson2019mimic} and apply histogram equalization using OpenCV \citep{opencv_library}, before storing the images in the JPEG format with a 95 quality factor. The code used to process the INSPECT volumes will be made available.

\paragraph{Models collected from}

\begin{itemize}
     \item \textbf{RAD-DINO:} Model and weights fetched from \href{https://huggingface.co/microsoft/rad-dino}{huggingface}.
    \item \textbf{GLoRIA:} Model collected from \href{https://github.com/marshuang80/gloria}{GitHub}, weights from
\href{https://stanfordmedicine.app.box.com/s/j5h7q99f3pfi7enc0dom73m4nsm6yzvh}{stanfordmedicine.app.box.com} (ResNet-50).
    \item \textbf{BioViL:} Model and weights downloaded through the HI-ML Multimodal Toolbox Python package \href{https://pypi.org/project/hi-ml-multimodal/}{pypi.org/project/hi-ml-multimodal/}.
    \item \textbf{MRM:} Acquired from \href{https://github.com/RL4M/MRM-pytorch}{https://github.com/RL4M/MRM-pytorch}.
    \item \textbf{BiomedCLIP:} Model and weights fetched from \href{https://huggingface.co/microsoft/BiomedCLIP-PubMedBERT_256-vit_base_patch16_224}{huggingface}.
    \item \textbf{C2L:} ResNet-18 weights downloaded from \href{https://github.com/funnyzhou/C2L_MICCAI2020}{GitHub}.
    \item \textbf{Medical MAE:} ViT-Base/16 weights (0.5M dataset) downloaded from \href{https://github.com/lambert-x/medical_mae}{GitHub}.
\end{itemize}

\paragraph{Model-specific preprocessing} Image preprocessing was chosen to match the preprocessing each backbone used during initial pre-training closely. All preprocessing not fetched from Huggingface was implemented using torchvision.

\begin{itemize}
    \item \textbf{RAD-DINO:} The preprocessor was fetched from the corresponding \href{https://huggingface.co/microsoft/rad-dino}{huggingface repository}.
    \item \textbf{GLoRIA \& Medical MAE:} Resized such that the shorter side $238$ pixels, followed by a $224 \times 224$ center crop. The pixel values were then rescaled to range $[0,1]$, and the three channels subsequently normalized according to the ImageNet mean and standard deviation (mean=$[0.485, 0.456, 0.406]$ and std=$[0.229, 0.224, 0.225]$).
    \item \textbf{BioViL:} Images were initially resized such the shorter side was $512$ pixels. $448 \times 448$ center-crop was applied followed by rescaling of values to range $[0,1]$
    \item \textbf{MRM:} Derived from \href{https://github.com/RL4M/MRM-pytorch}{https://github.com/RL4M/MRM-pytorch}. Resize to $224$ pixels followed by $224 \times 224$ center crop. The image is converted to grayscale, rescaled to range $[0,1]$ and normalized with mean=$0.4978$ and std=$0.2449$.
    \item \textbf{BiomedCLIP:} Fetched from \href{https://huggingface.co/microsoft/BiomedCLIP-PubMedBERT_256-vit_base_patch16_224}{huggingface}.
    \item \textbf{C2L:} Derived from \href{https://github.com/funnyzhou/C2L_MICCAI2020}{GitHub}. Image resize to $224$ pixels followed by a $224 \times 224$ center-crop. Channels are rescaled to range $[0,1]$ and normalized according to the ImageNet mean and std (provided previously).
\end{itemize}

For data augmentation, we used random resized crop between scales $0.4$ and $0.9$. During training, layer normalization was applied after extracting the pre-trained encoder features (\ie, applied to $V(x)$ and/or $T(z)$). An additional normalization was used for the teacher model before concatenating the (self-attended and mean-pooled) image and text representations. Self-attention layers used a dropout layer ($p = 0.2$) on the attention weights before multiplying them with the value vector.

\paragraph{Training sizes and epochs} Tables~\ref{tab:epochs1}--\ref{tab:epochs5} cover the training dataset sizes, and their corresponding number of training epochs, used for the experiments in the main paper. Epochs were chosen such that all models had time to converge. We additionally plan to release the specific train-test splits for each seed.

\begin{table}[H]
  \centering
  \begin{minipage}[t]{0.30\textwidth}
    \centering
    \caption{\textbf{MIMIC-CXR-JPG}}
    \label{tab:epochs1}
    \begin{tabular}{lcc}
      \toprule
      \small{Fraction} & \small{\#samples} & \small{Epochs}\\
      \midrule
      $2.5\%$ & $100$ & $100$\\
      $5\%$ & $200$ & $100$\\
      $10\%$ & $400$ & $100$\\
      $25\%$ & $1000$ & $50$\\
      $50\%$ & $2000$ & $50$\\
      $100\%$ & $4000$ & $50$\\
      \bottomrule
    \end{tabular}
  \end{minipage}
  \hfill
  \begin{minipage}[t]{0.30\textwidth}
    \centering
    \caption{\textbf{Age}}
    \label{tab:epochs2}
    \begin{tabular}{lcc}
      \toprule
      \small{Fraction} & \small{\#samples} & \small{Epochs}\\
      \midrule
      $1\%$ & $317$ & $150$\\
      $2.5\%$ & $792$ & $100$\\
      $5\%$ & $1585$ & $100$\\
      $10\%$ & $3171$ & $50$\\
      \bottomrule
    \end{tabular}
  \end{minipage}
  \hfill
  \begin{minipage}[t]{0.30\textwidth}
    \centering
    \caption{\textbf{3-Day Discharge}}
    \label{tab:epochs3}
    \begin{tabular}{lcc}
      \toprule
      \small{Fraction} & \small{\#samples} & \small{Epochs}\\
      \midrule
      $0.5\%$ & $82$ & $100$\\
      $1\%$ & $165$ & $100$\\
      $2.5\%$ & $416$ & $100$\\
      $5\%$ & $829$ & $50$\\
      $10\%$ & $1662$ & $50$\\
      $25\%$ & $4145$ & $50$\\
      \bottomrule
    \end{tabular}
  \end{minipage}

    \centering
  \begin{minipage}[t]{0.30\textwidth}
    \centering
    \caption{\textbf{Readmission}}
    \label{tab:epochs4}
    \begin{tabular}{lcc}
      \toprule
      \small{Fraction} & \small{\#samples} & \small{Epochs}\\
      \midrule
      $5\%$ & $211$ & $100$\\
      $10\%$ & $423$ & $100$\\
      $17.5\%$ & $741$ & $100$\\
      $25\%$ & $1059$ & $100$\\
      $50\%$ & $2119$ & $50$\\
      \bottomrule
    \end{tabular}
  \end{minipage}
  \qquad
  \begin{minipage}[t]{0.30\textwidth}
    \centering
    \caption{\textbf{12-Month PH}}
    \label{tab:epochs5}
    \begin{tabular}{lcc}
      \toprule
      \small{Fraction} & \small{\#samples} & \small{Epochs}\\
      \midrule
      $5\%$ & $204$ & $100$\\
      $10\%$ & $408$ & $100$\\
      $17.5\%$ & $715$ & $100$\\
      $25\%$ & $1022$ & $100$\\
      $50\%$ & $2044$ & $50$\\
      \bottomrule
    \end{tabular}
  \end{minipage}
\end{table}

\paragraph{Pre-training datasets} Table \ref{tab:pretrain_datasets} offers an overview of the datasets each model has been pre-trained with.

\begin{table*}[h]
  \caption{An overview of the pre-training datasets of the models compared in this study.}
  \label{tab:pretrain_datasets}
  \centering
  \begin{tabular}{llcccccc}
    \toprule
    & \multicolumn{4}{c}{Pre-Trained On}\\
     \cmidrule(r){2-5}
    Model & \small{MIMIC} & \small{CheXpert} & NIH-CXR & \small{Other}\\
    \midrule
    RAD-DINO & \cmark & \cmark & \cmark & \cmark   \\
    C2L & \cmark & \cmark & \cmark & \cmark \\
    Medical MAE & \cmark & \cmark & \cmark & \xmark \\
    MRM  & \cmark & \xmark & \xmark & \xmark \\
    BioViL-T & \cmark & \xmark & \xmark & \xmark\\
    GLoRIA & \xmark & \cmark & \xmark & \xmark\\
    BiomedCLIP & \xmark & \xmark & \xmark & \cmark \\
    \bottomrule
  \end{tabular}
\end{table*}

\newpage
\section{Extended Results}
\label{sec:ext_results}

\subsection{AUPRC Figures}

\begin{figure*}[h]
\centering
     \resizebox{0.85\textwidth}{!}{
        \begin{minipage}{\textwidth}

    \begin{subfigure}[T]{0.33\textwidth}
        \centering
        \includegraphics[width=0.99\linewidth]{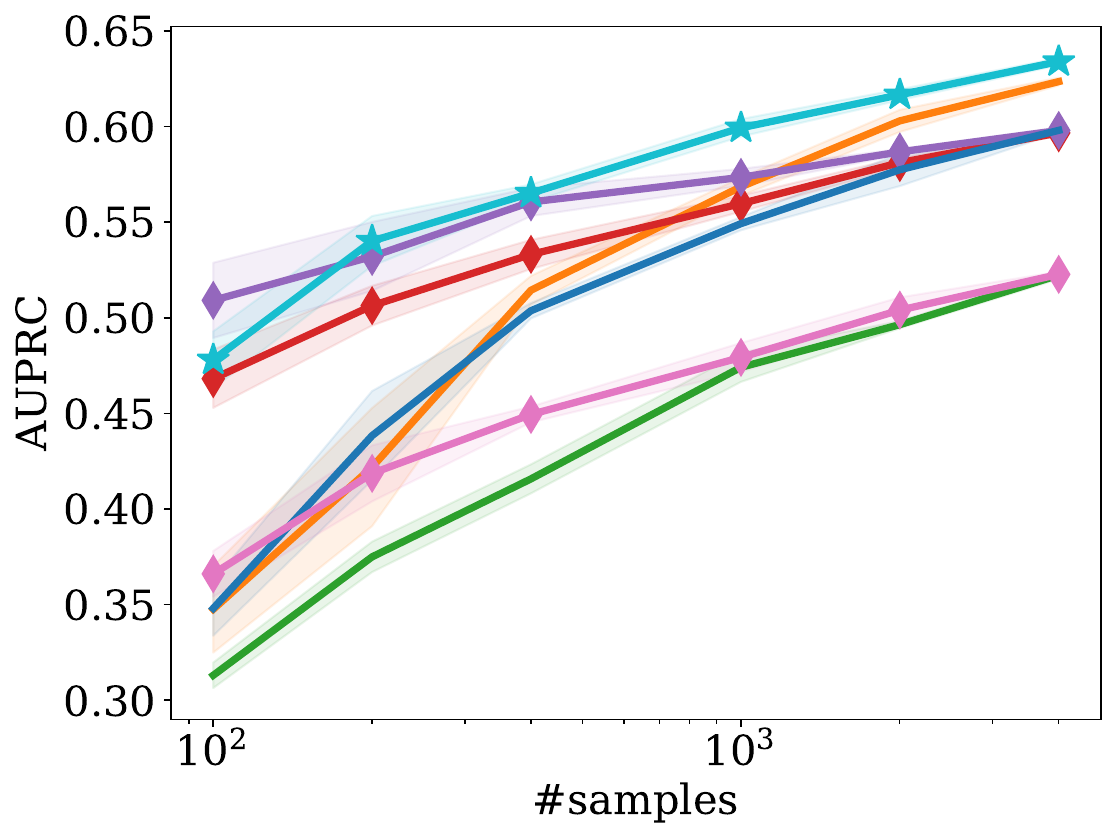}
        \caption{\textbf{MIMIC 5x1200}}
    \end{subfigure}
    \hfill
    \begin{subfigure}[T]{0.33\textwidth}
       \centering
        \includegraphics[width=0.99\linewidth]{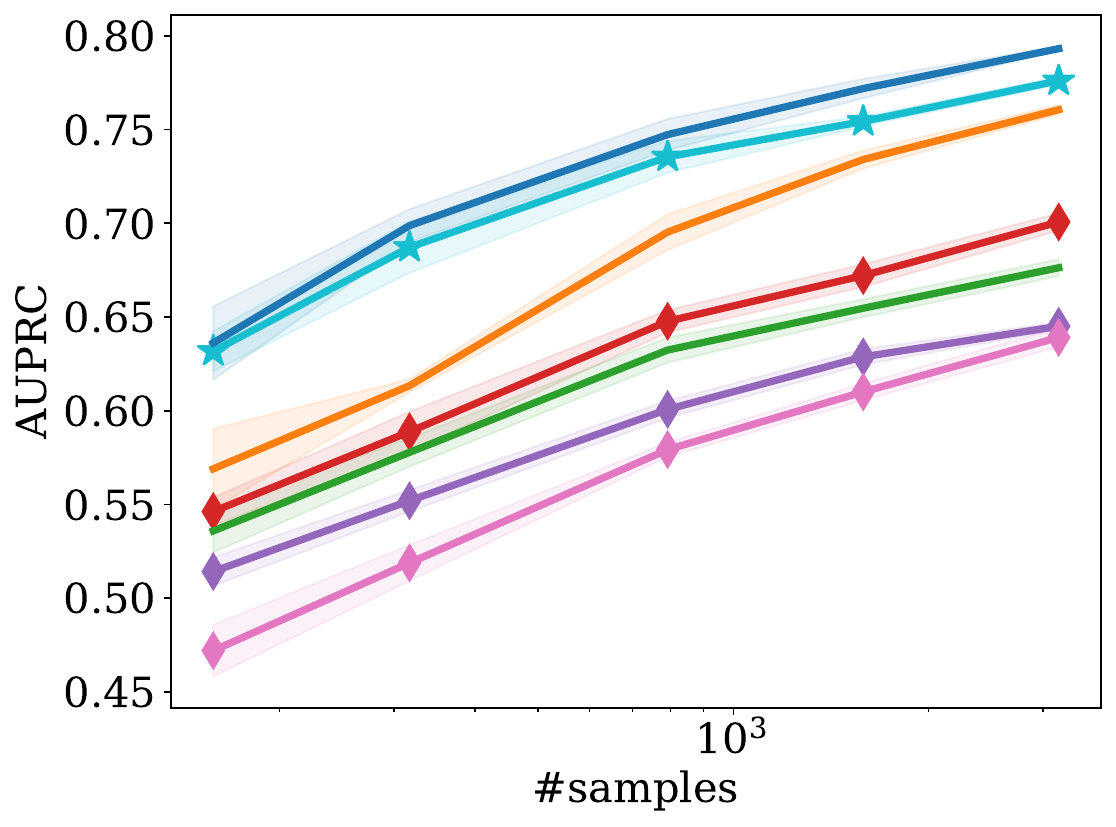}
        \caption{\textbf{Age}}
    \end{subfigure}
    \centering
    \begin{subfigure}[T]{0.33\textwidth}
        \centering
        \includegraphics[width=0.99\linewidth]{plots/pretraining_comp/with_mae/legend_combined.pdf}
    \end{subfigure}
    \hfill
    \begin{subfigure}[T]{0.33\textwidth}
        \centering
         \includegraphics[width=0.99\linewidth]{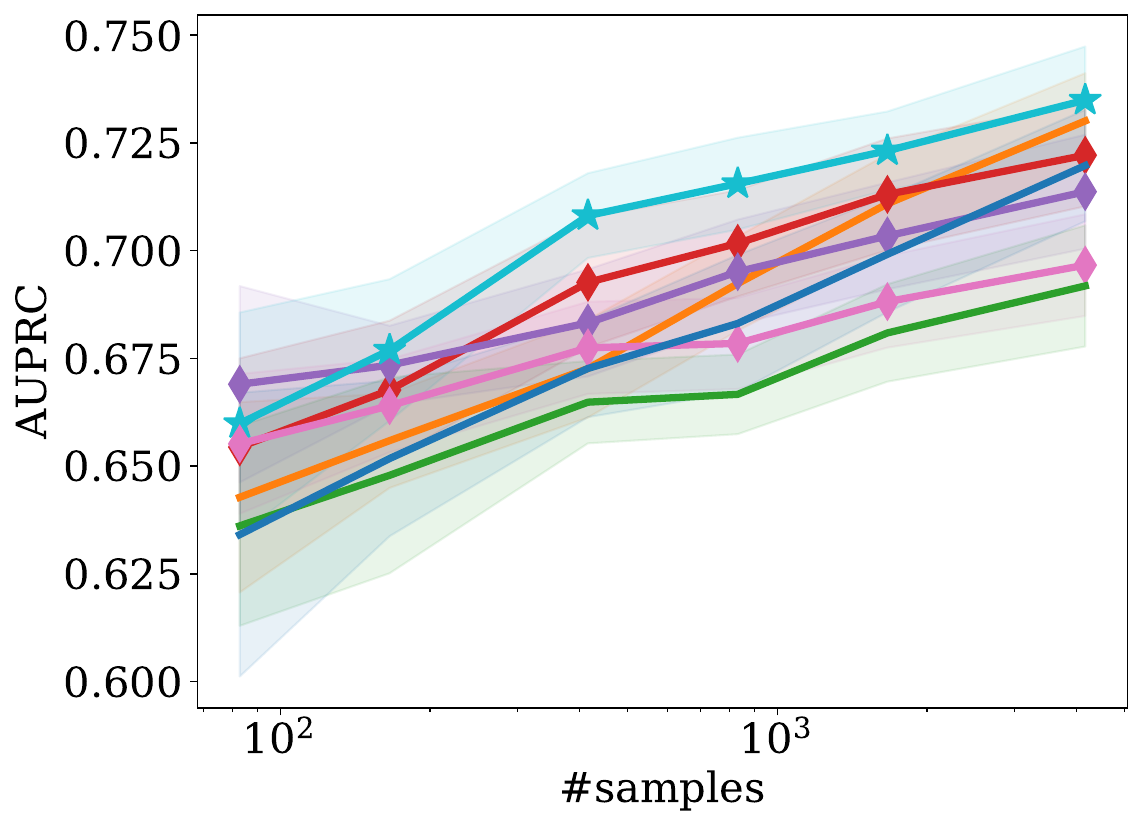}
        \caption{\textbf{3-Day Discharge}}
    \end{subfigure}
    \hfill
    \begin{subfigure}[T]{0.33\textwidth}
       \centering
        \includegraphics[width=0.99\linewidth]{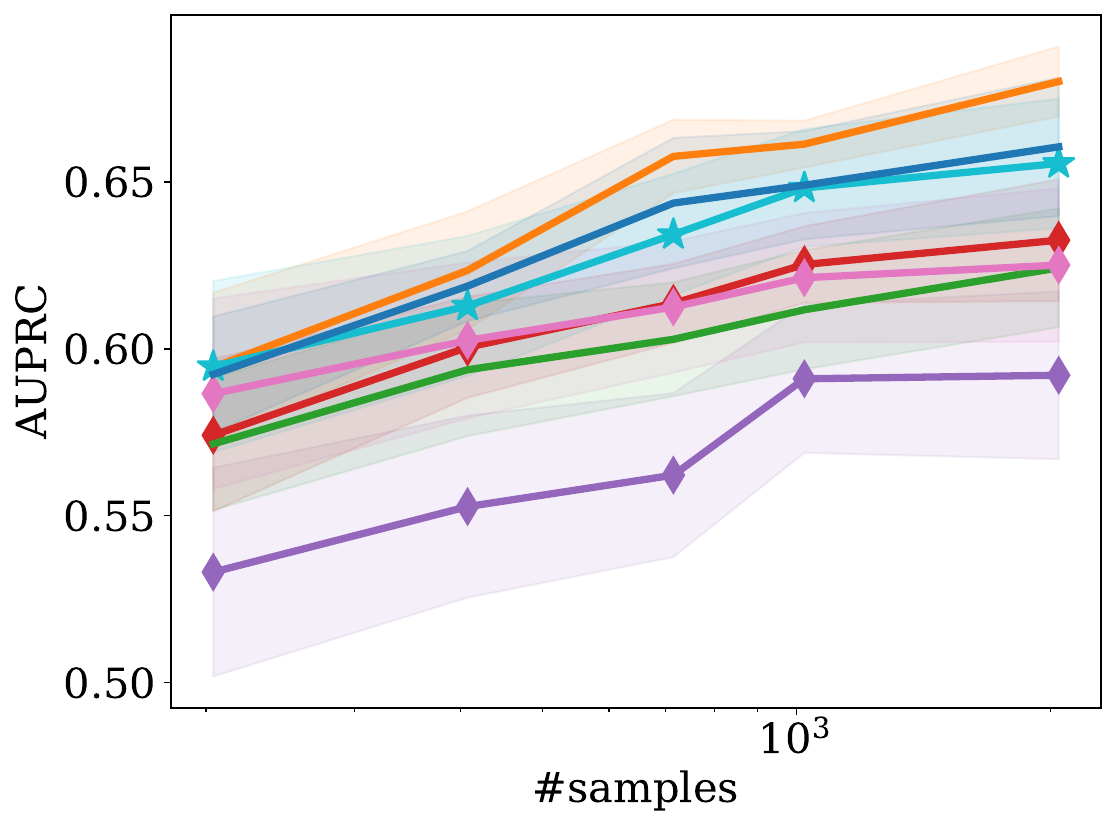}
        \caption{\textbf{12-Month PH}}
    \end{subfigure}\hfill
    \begin{subfigure}[T]{0.33\textwidth}
       \centering
        \includegraphics[width=0.99\linewidth]{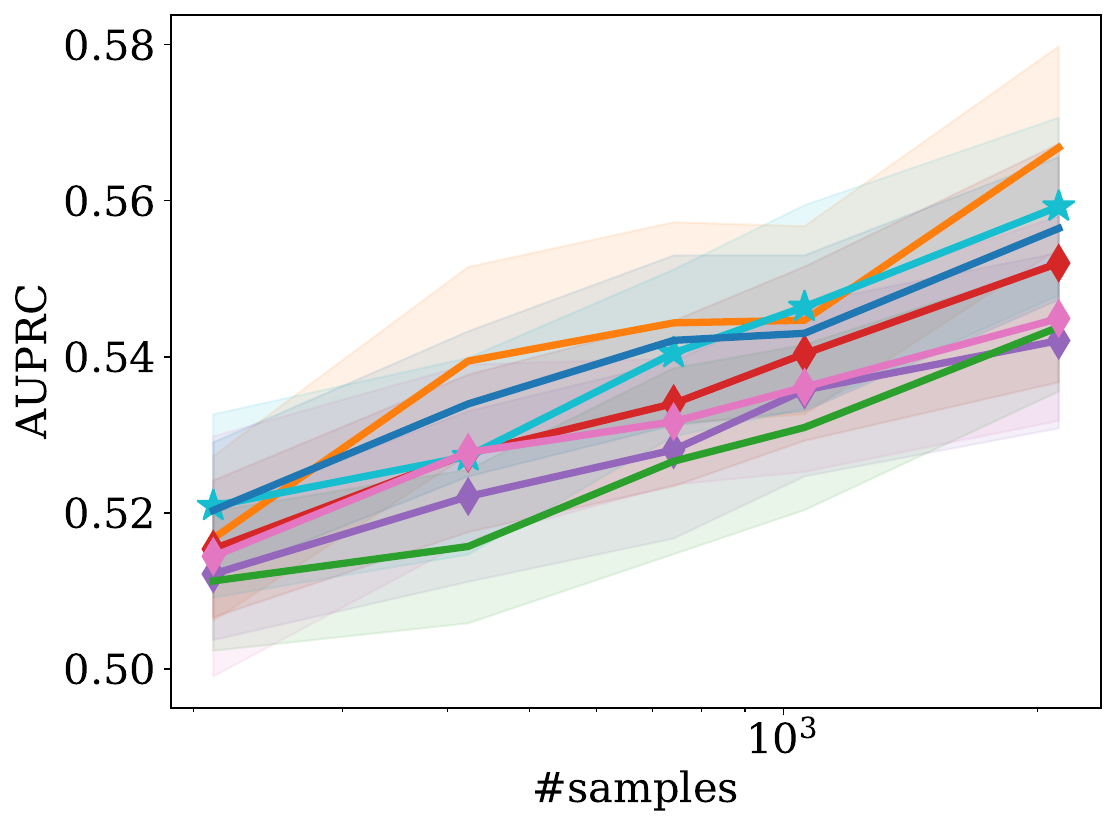}
        \caption{\textbf{Readmission}}
    \end{subfigure}
    \end{minipage}
    }
    \caption{A comparison of the sample efficiency of different backbones. The plots are averaged over 5 seeds, except for Readmission, which is averaged over 10. The shaded area regions represent the $90\%$ CI.} 
    \label{fig:backbone_comp_ap}
\end{figure*}

\begin{figure*}[!h]
    \centering
     \resizebox{0.84\textwidth}{!}{
        \begin{minipage}{\textwidth}

    \centering
    \includegraphics[width=0.99\textwidth]{plots/bar_legend_only_mae.pdf}
    
    \centering
    \begin{subfigure}[B]{0.64\textwidth}
        \centering
        \includegraphics[width=0.54\textwidth]{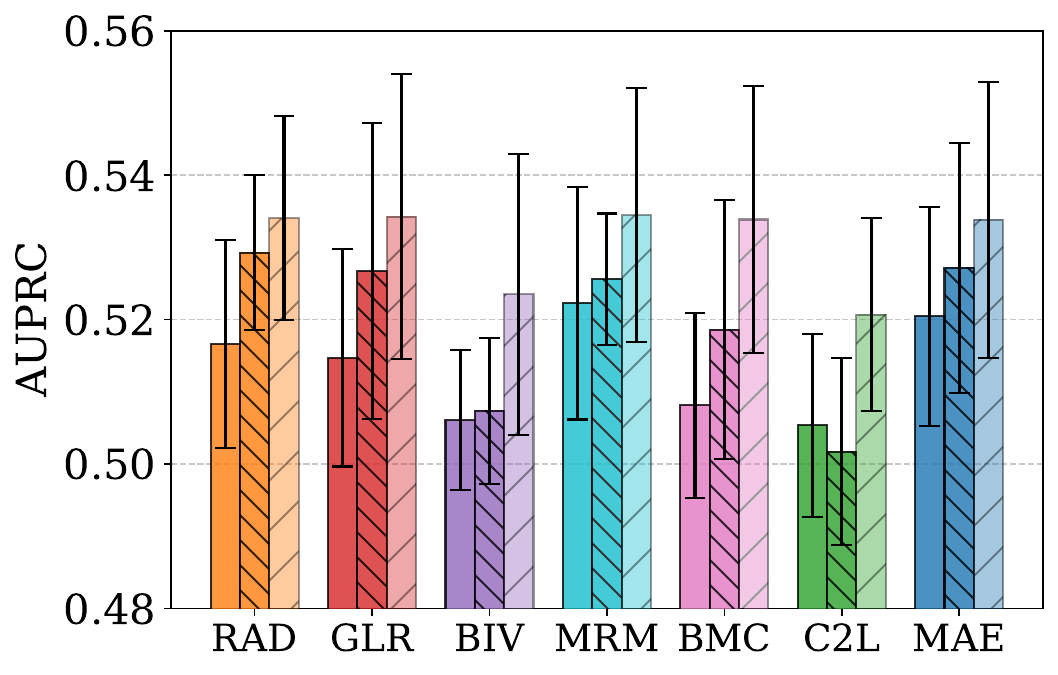}
        \includegraphics[width=0.45\textwidth]{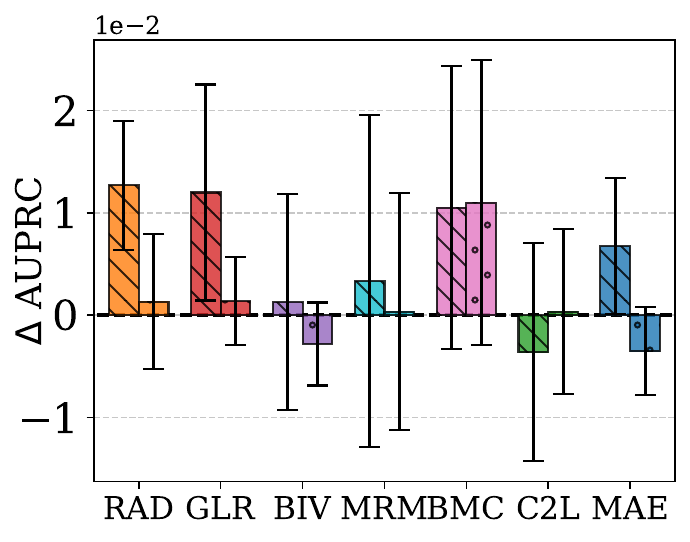}
        \caption{$5 \%$ of \textbf{Readmission} (211 samples) Left: Average AUPRC \\Right: Average AUPRC-difference  between distillation and image-only}
    \end{subfigure}\hfill
    \hfill
    \begin{subfigure}[B]{0.345\textwidth}
        \centering\includegraphics[width=\textwidth]{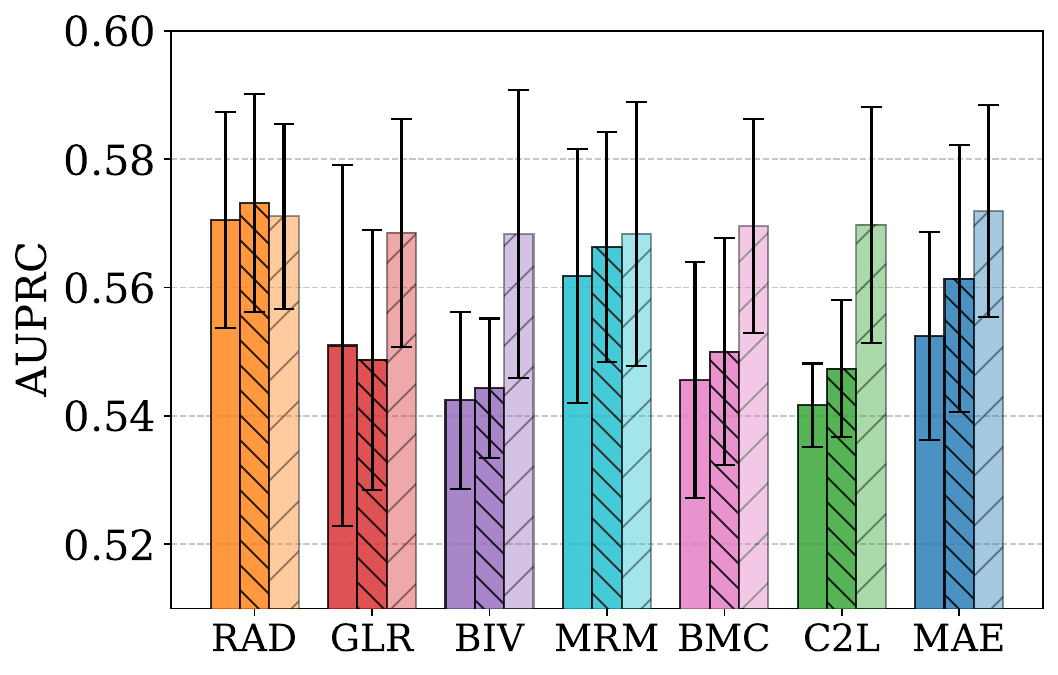}
        \caption{$50 \%$ of \textbf{Readmission} (2119 samples) Average AUPRC}
    \end{subfigure}
    \centering
    \centering
    \begin{subfigure}[B]{0.64\textwidth}
        \centering
        \includegraphics[width=0.54\textwidth]{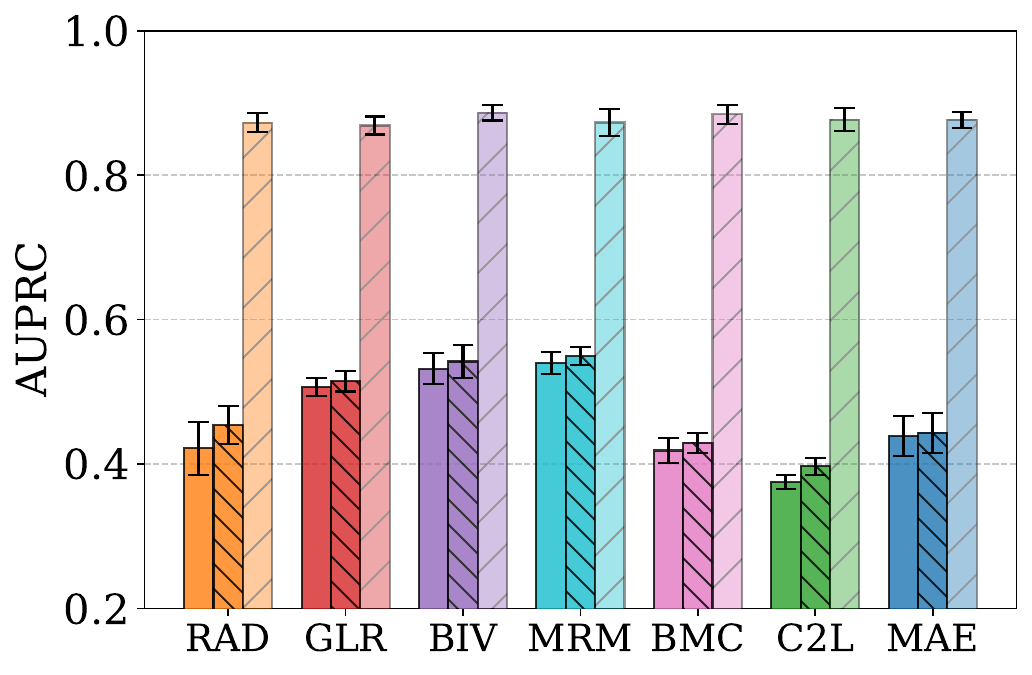}
        \includegraphics[width=0.45\textwidth]{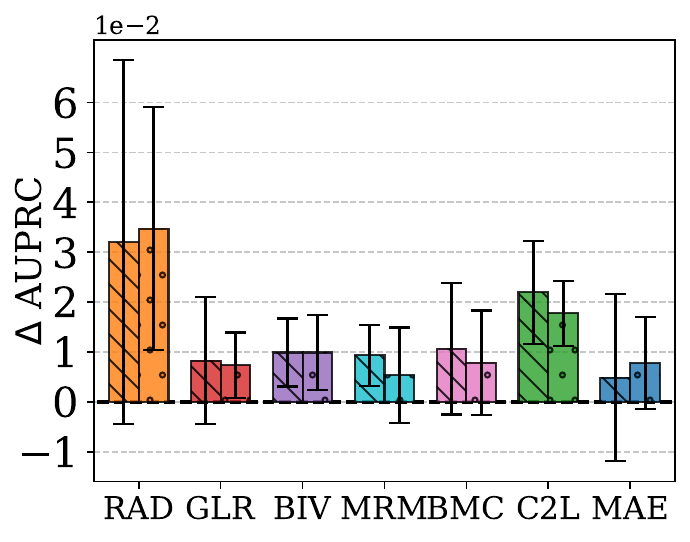}
        \caption{$5 \%$ of \textbf{MIMIC 5x1200} (200 samples) Left: Average AUPRC \\Right: Average AUPRC-difference  between distillation and image-only}
    \end{subfigure}\hfill
    \hfill
    \begin{subfigure}[B]{0.345\textwidth}
        \centering\includegraphics[width=\textwidth]{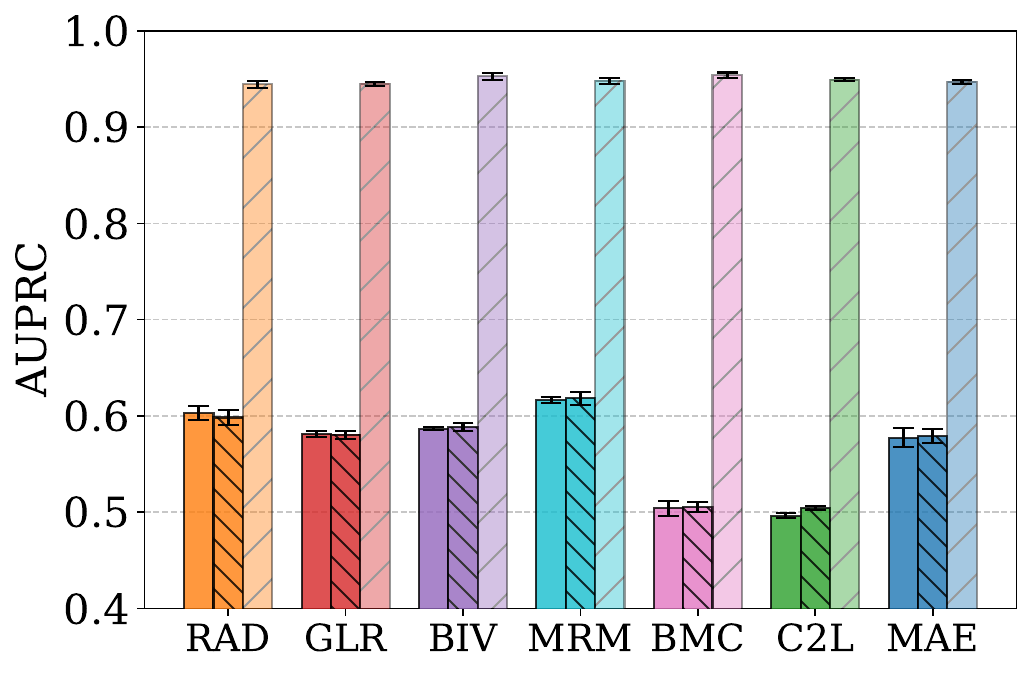}
        \caption{$50 \%$ of \textbf{MIMIC 5x1200} (2000 samples) Average AUPRC}
    \end{subfigure}
    \end{minipage}
    }
    \caption{Distillation results on \textbf{Readmission} (prognostic) and \textbf{MIMIC 5x1200} (diagnostic) with different training set sizes, averaged over 5 seeds. Error bars represent the $95\%$ confidence intervals.}
    \label{fig:barplot_ap}
\end{figure*}

Figure \ref{fig:backbone_comp_ap} shows the AUPRC for the experiments in Figure \ref{fig:backbone_comp}. Micro-averaging was used for the multi-class tasks. We provide the AUPRC for the distillation experiments in Figure \ref{fig:barplot_ap}.
\subsection{Diagnostic Label INSPECT}

\begin{figure}[H]

\centering
    \includegraphics[width=0.99\textwidth]{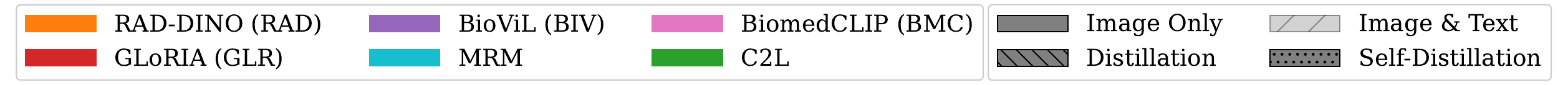}
    
    \centering
    \begin{subfigure}[B]{0.64\textwidth}
        \centering
        \includegraphics[width=0.54\textwidth]{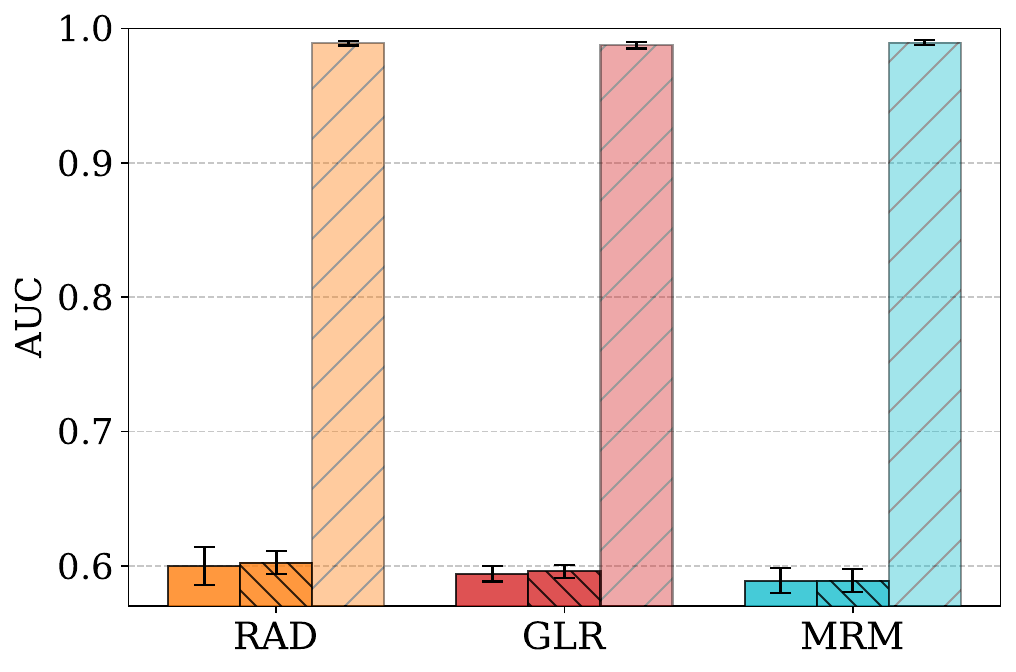}
        \includegraphics[width=0.45\textwidth]{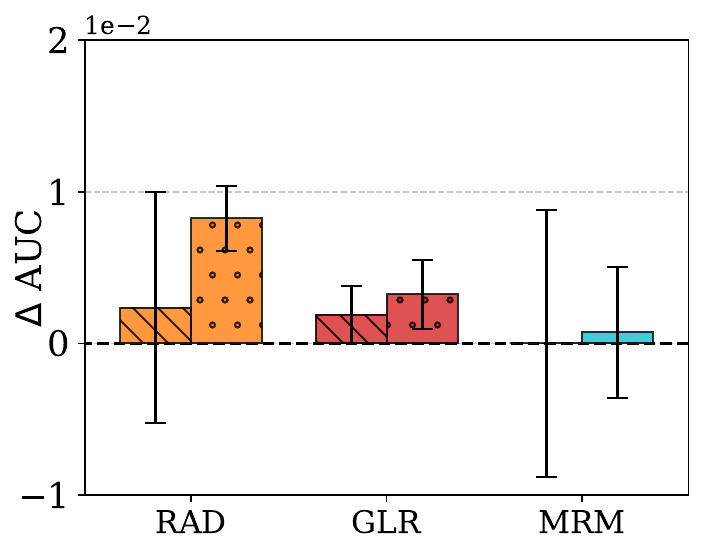}
        \caption{$5 \%$ of training data (351 samples) Left: Average AUC \\Right: Average AUC-difference  between distillation and image-only}
        \label{fig:pe_0.05}
    \end{subfigure}\hfill
    \hfill
    \begin{subfigure}[B]{0.345\textwidth}
        \centering\includegraphics[width=\textwidth]{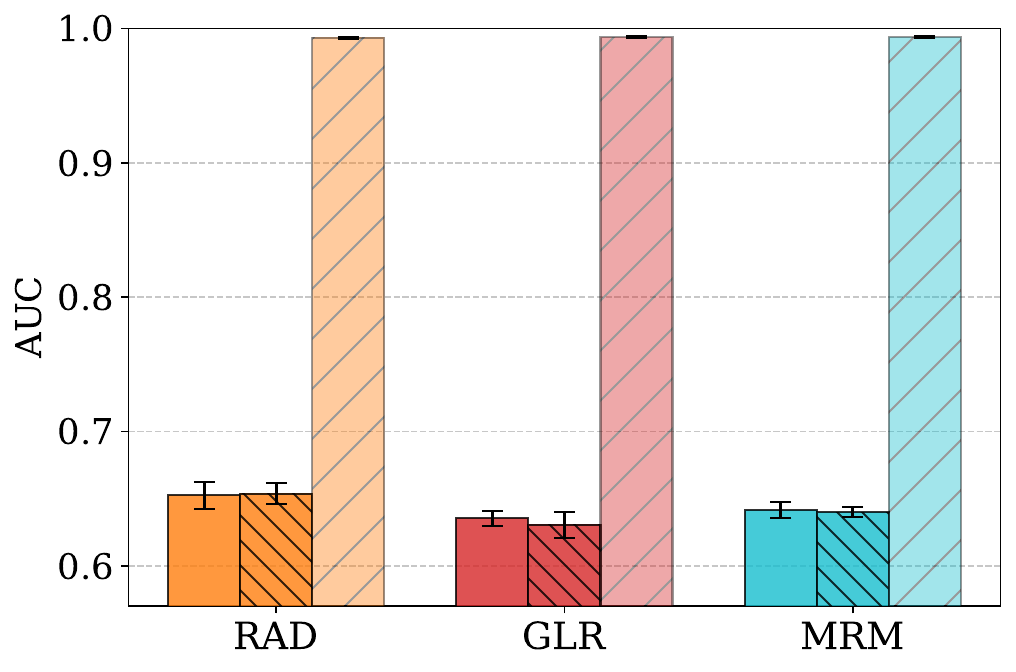}
        \caption{$50 \%$ of training data (3515 samples) Average AUC}
        \label{fig:pe_0.5}
    \end{subfigure}
    \caption{\textbf{Pulmonary Embolism}}
    \label{fig:pe}
\end{figure}

We perform an additional experiment on the INSPECT dataset, in which models are trained to predict whether or not a patient currently suffers from \textbf{Pulmonary Embolism}. As before, this diagnostic label has been extracted from the accompanying radiology reports. Similar to \textbf{Readmission} and \textbf{12-Month PH}, we artificially balance the dataset by sub-sampling the number of negative labels to match the number of positive. The final dataset consists of $9,375$ images, where $75\%$ are designated for training and $25\%$ for validation. The results (Figure \ref{fig:pe}) again indicate poor distillation performance when the text is too predictive of the label.

\subsection{Age and 12-Month PH results}

\begin{figure}[H]
    \centering
    \includegraphics[width=0.99\textwidth]{plots/bar_legend_only_mae.pdf}

\centering
    \begin{subfigure}[T]{0.44\textwidth}
       \centering
        \includegraphics[width=0.95\linewidth]{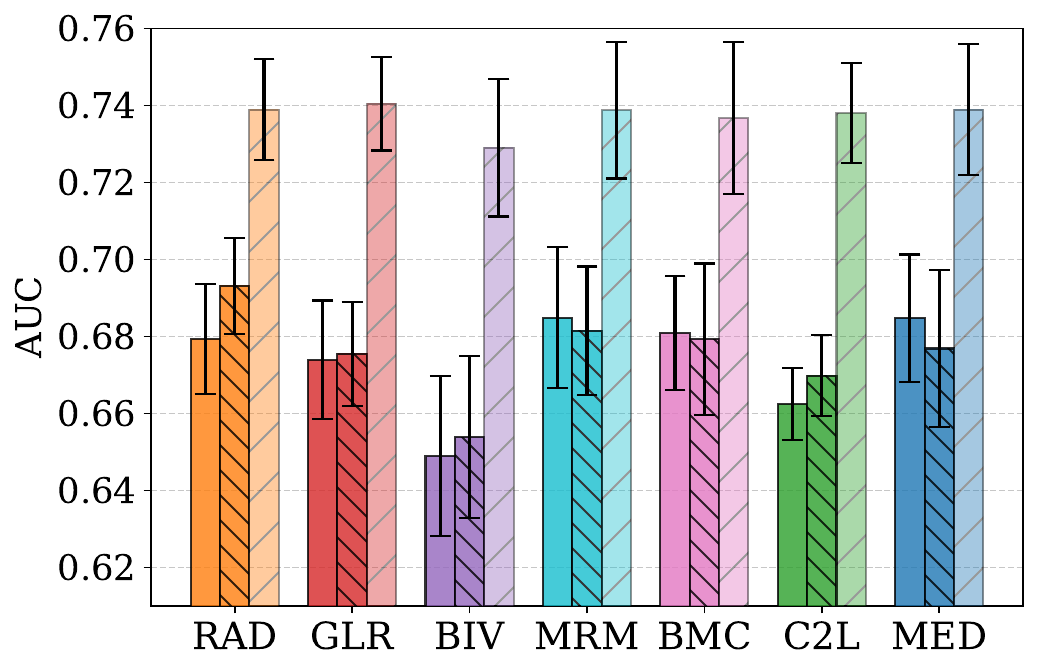}
        \caption{$5 \%$ of training data (204 samples)}
    \end{subfigure}
    \begin{subfigure}[T]{0.44\textwidth}
        \centering
        \includegraphics[width=0.95\linewidth]{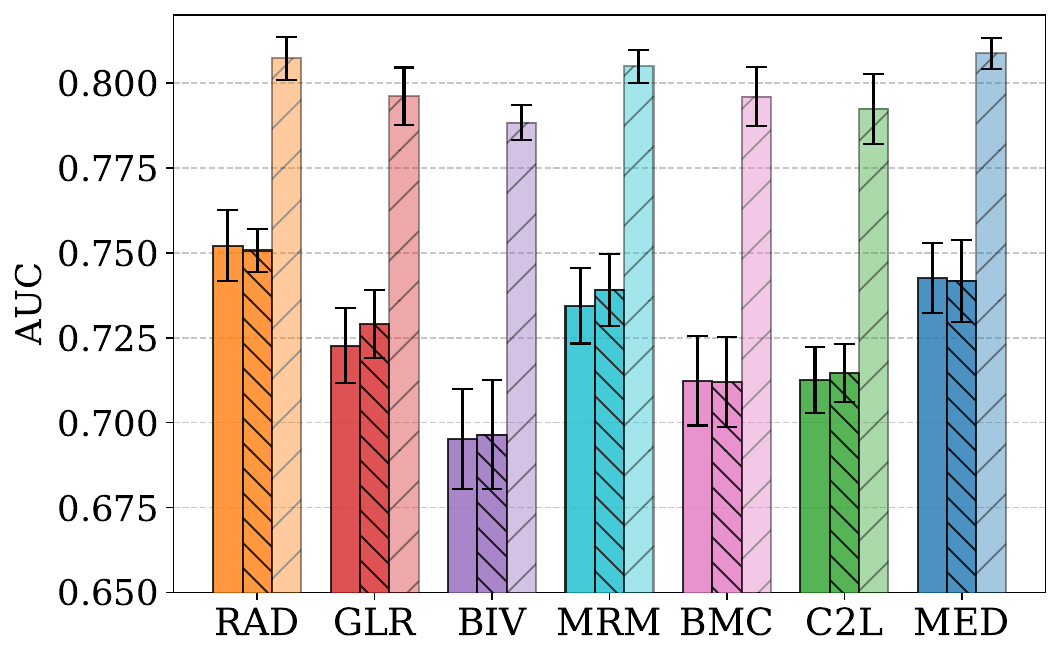}
        \caption{$50 \%$ of training data (2044 samples)}
    \end{subfigure}
    \caption{Mean AUC, averaged over 5 seeds, when performing distillation on \textbf{12-Month PH}.}
    \label{fig:12_month_app}
\end{figure}

\begin{figure}[H]
\centering
    \begin{subfigure}[T]{0.44\textwidth}
       \centering
        \includegraphics[width=0.95\linewidth]{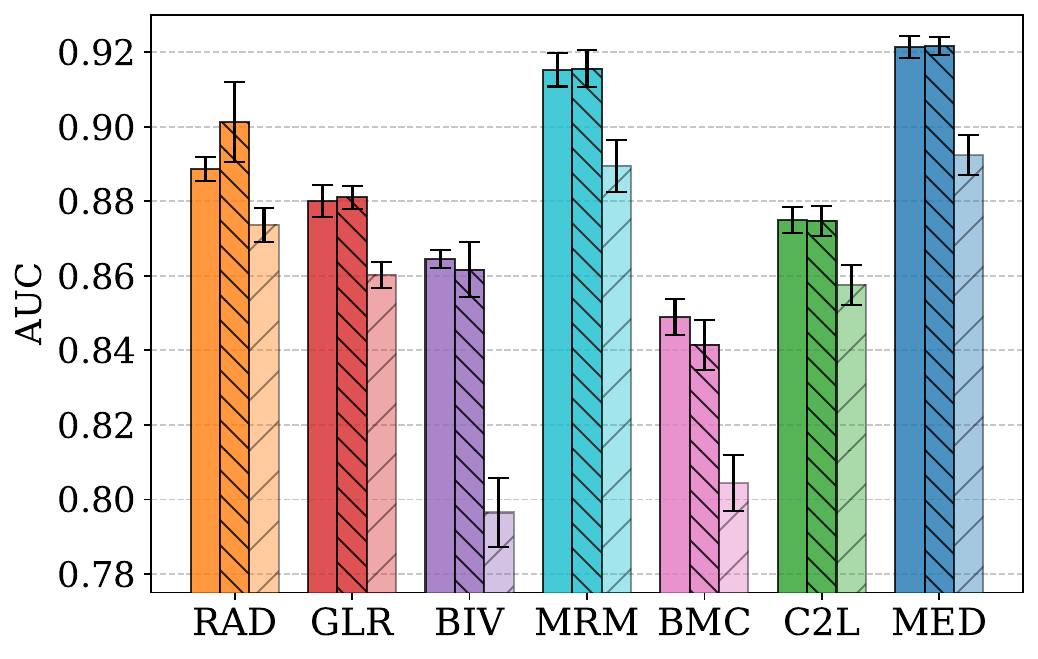}
        \caption{$1 \%$ of training data (317 samples)}
    \end{subfigure}
    \begin{subfigure}[T]{0.44\textwidth}
        \centering
        \includegraphics[width=0.95\linewidth]{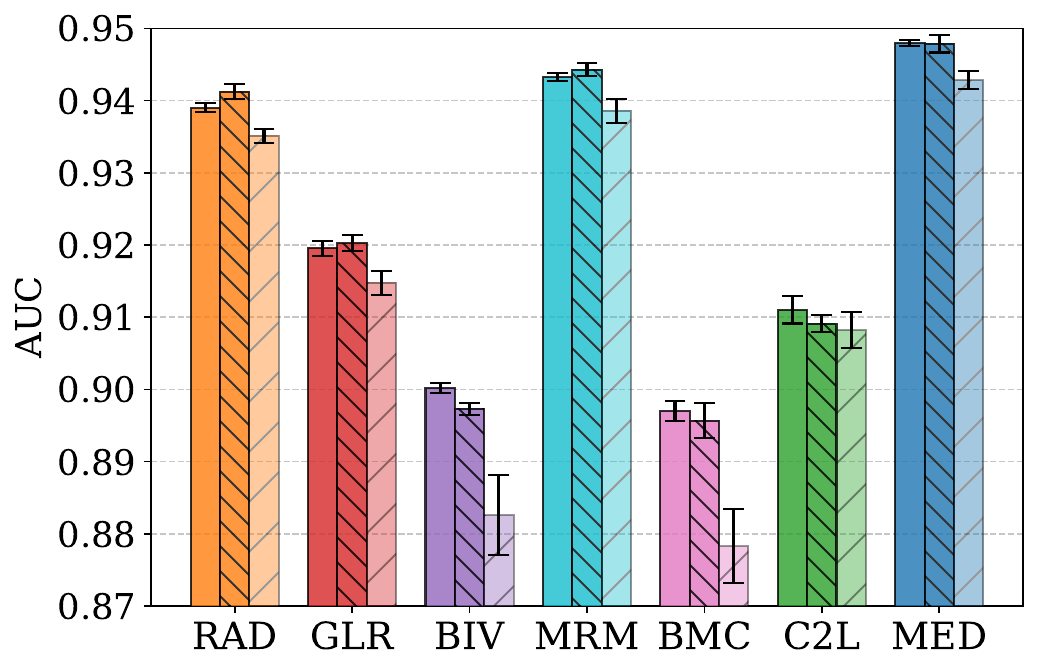}
        \caption{$10 \%$ of training data (3171 samples)}
    \end{subfigure}
    \caption{Mean AUC, averaged over 5 seeds, when performing distillation on \textbf{Age}.}
    \label{fig:age_app}
\end{figure}

We include the distillation results on the \textbf{12-Month PH} (Figure \ref{fig:12_month_app}) and \textbf{Age} (Figure \ref{fig:age_app}) datasets. Notably, in the Age experiment, the teacher performs worse on average for every backbone (even with more than $3,000$ training samples), suggesting that the image is substantially more predictive than the radiology reports.

\subsection{Dino \& MRM Seeds}

\begin{figure}[H]
    \centering
    \begin{subfigure}{0.46\textwidth}
        \centering
        \includegraphics[width=\textwidth]{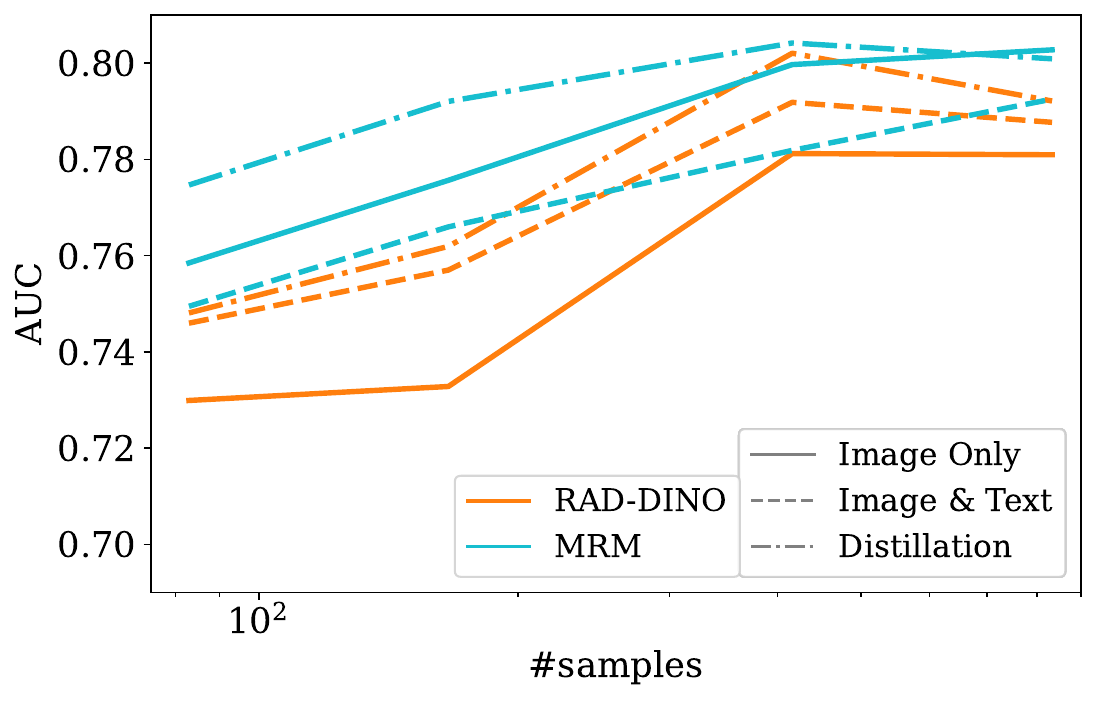}
        \caption{Seed 0}
    \end{subfigure}\hfill
    \begin{subfigure}{0.46\textwidth}
        \centering
        \includegraphics[width=\textwidth]{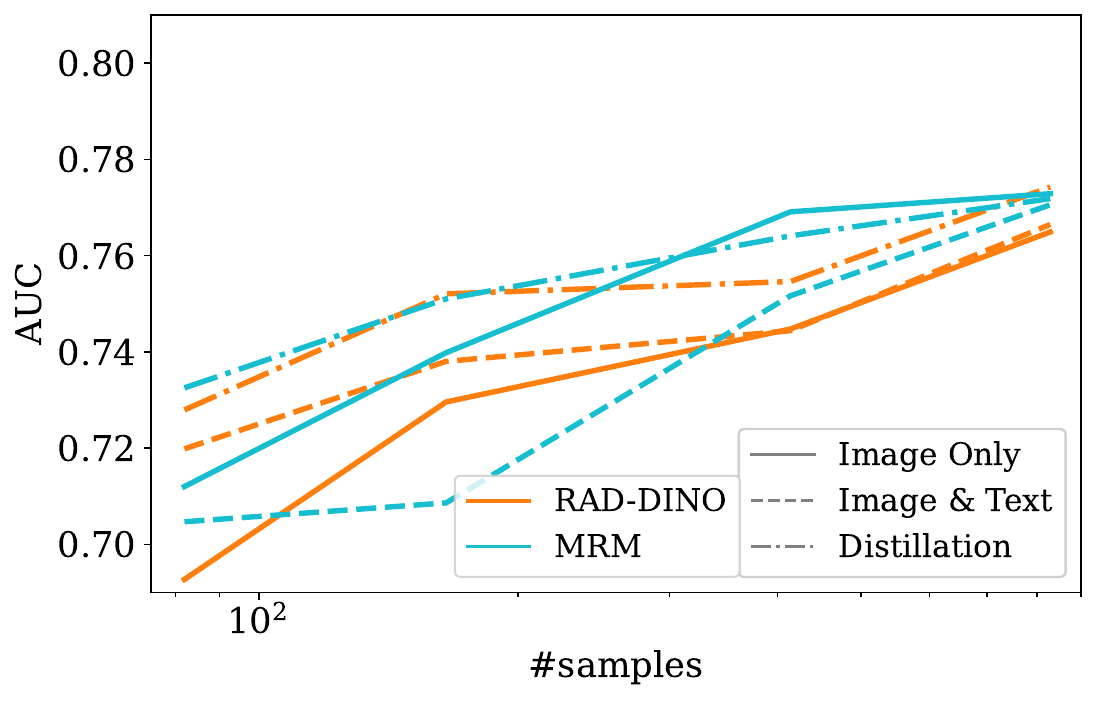}
        \caption{Seed 1}
    \end{subfigure}
    \centering
    \begin{subfigure}{0.46\textwidth}
        \centering
        \includegraphics[width=\textwidth]{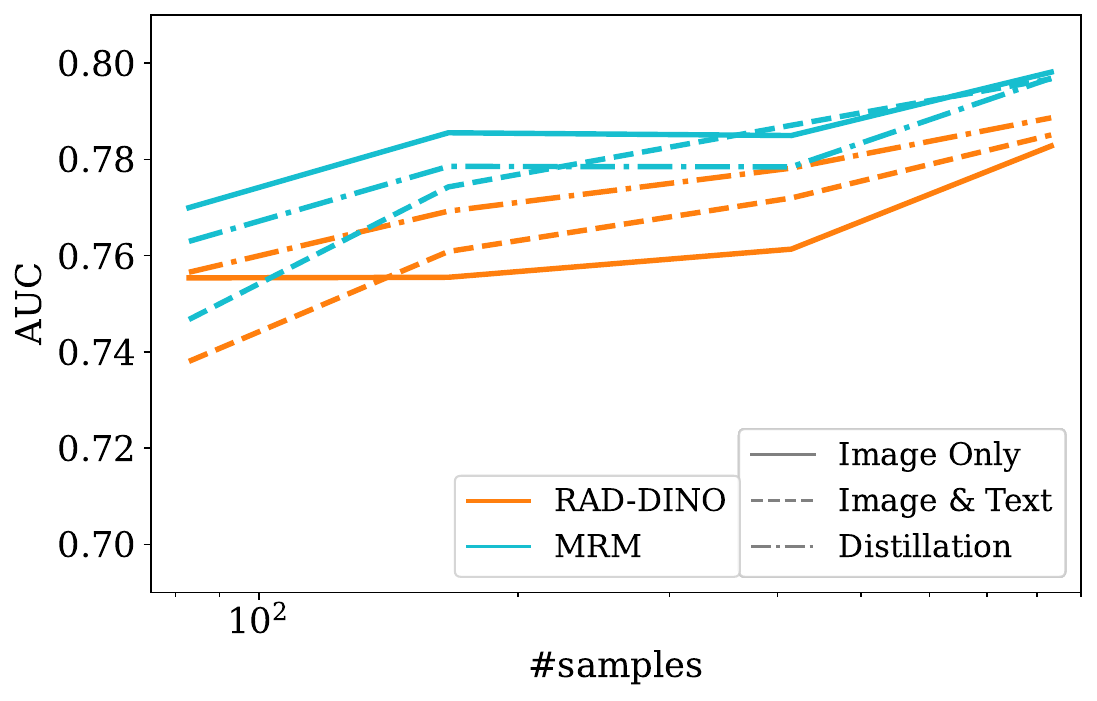}
        \caption{Seed 2}
    \end{subfigure}\hfill
    \begin{subfigure}{0.46\textwidth}
        \centering
        \includegraphics[width=\textwidth]{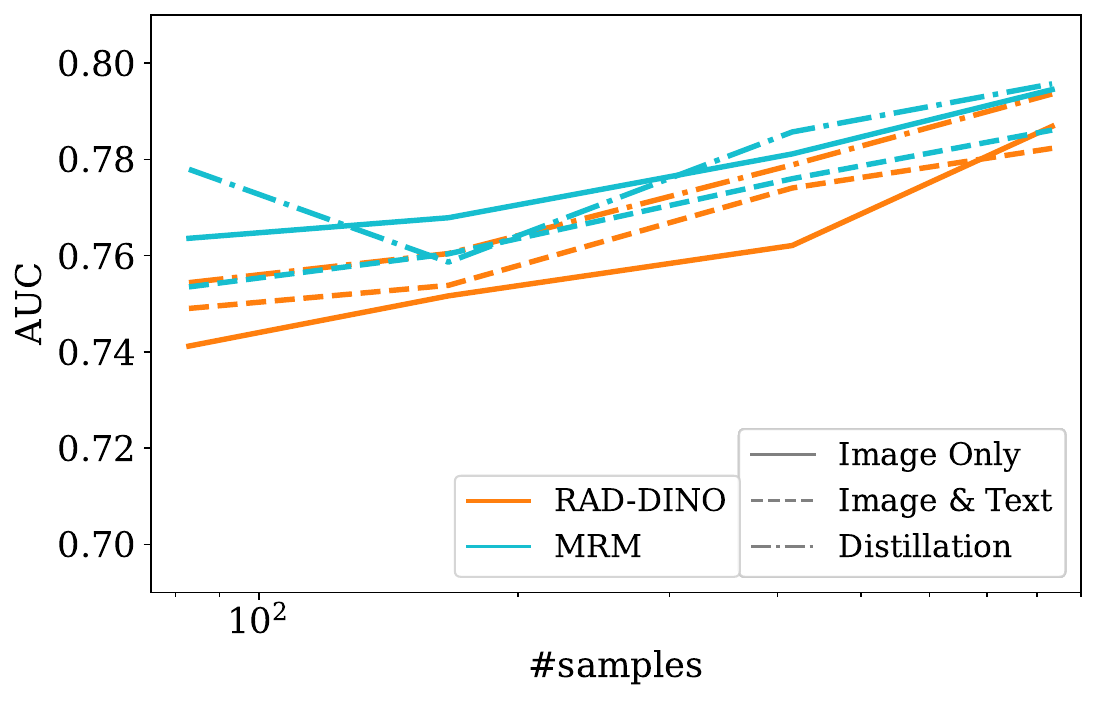}
        \caption{Seed 3}
    \end{subfigure}
    \centering
    \begin{subfigure}{0.46\textwidth}
        \centering
        \includegraphics[width=\textwidth]{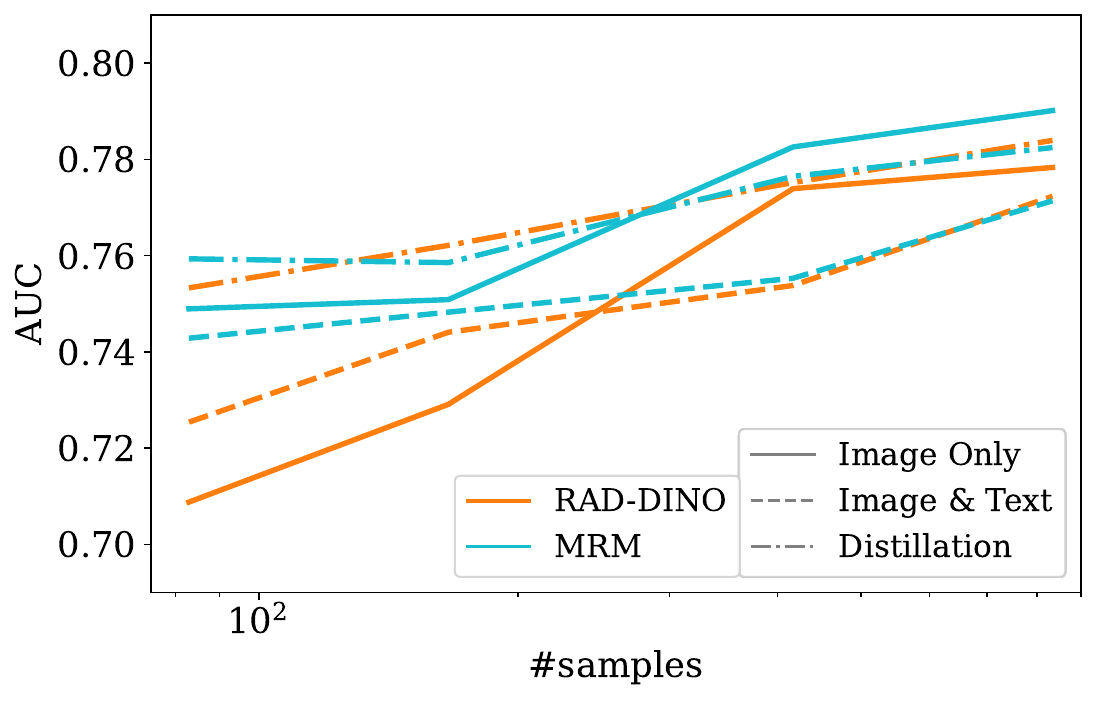}
        \caption{Seed 4}
    \end{subfigure}
    \caption{\textbf{3-Day Discharge} - RAD-DINO and MRM performance across the 5 seeds averaged in Figure \ref{fig:discharge_dino}. In the case of RAD-DINO, the student consistently outperforms the teacher.}
    \label{fig:dino_seeds}
\end{figure}

\newpage
\subsection{Readmission and MIMIC 5x1200 Seeds}

\begin{figure}[H]
    \centering
    \begin{subfigure}{0.46\textwidth}
        \centering
        \includegraphics[width=\textwidth]{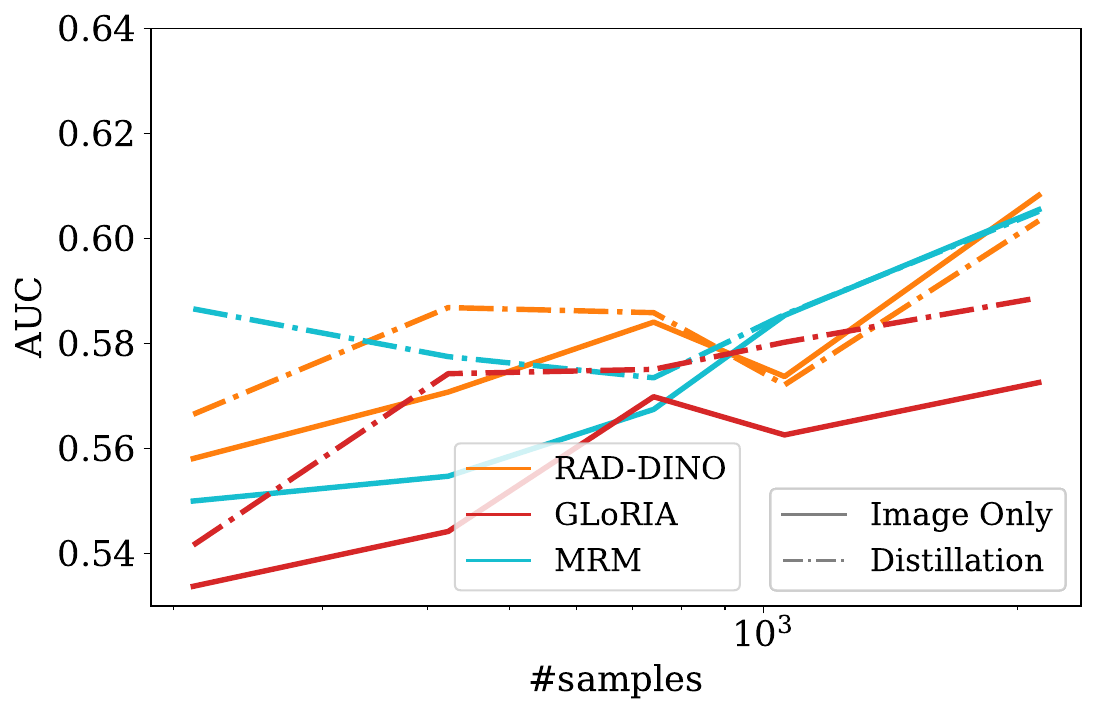}
        \caption{Seed 0}
    \end{subfigure}\hfill
    \begin{subfigure}{0.46\textwidth}
        \centering
        \includegraphics[width=\textwidth]{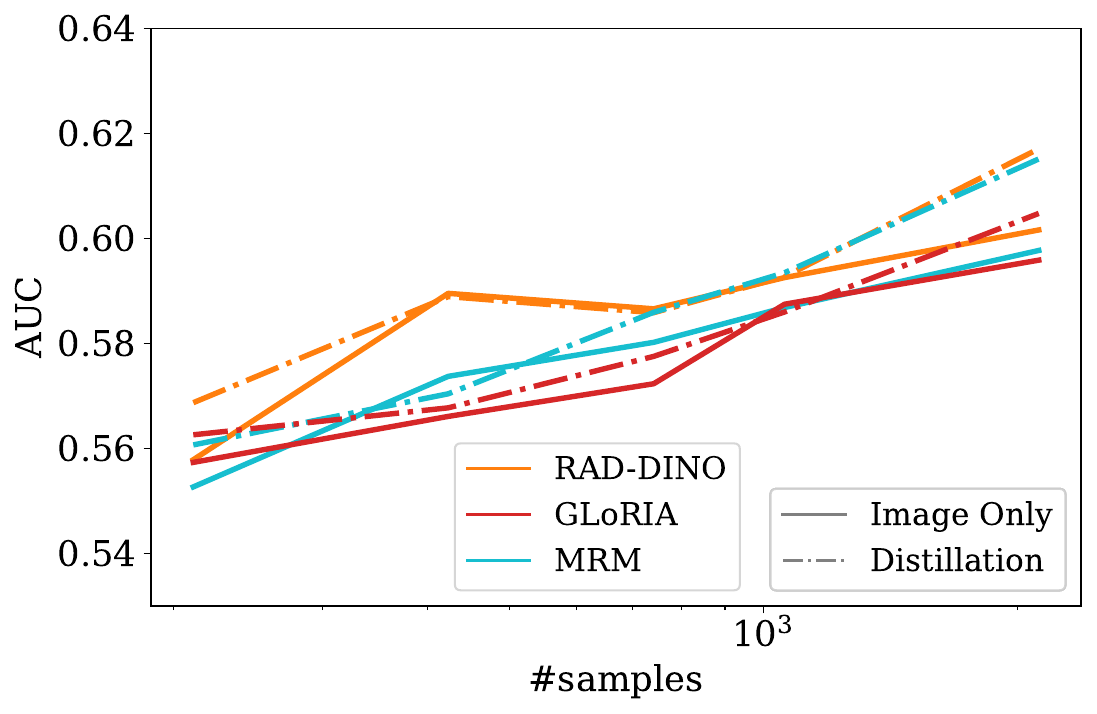}
        \caption{Seed 1}
    \end{subfigure}
    \centering
    \begin{subfigure}{0.46\textwidth}
        \centering
        \includegraphics[width=\textwidth]{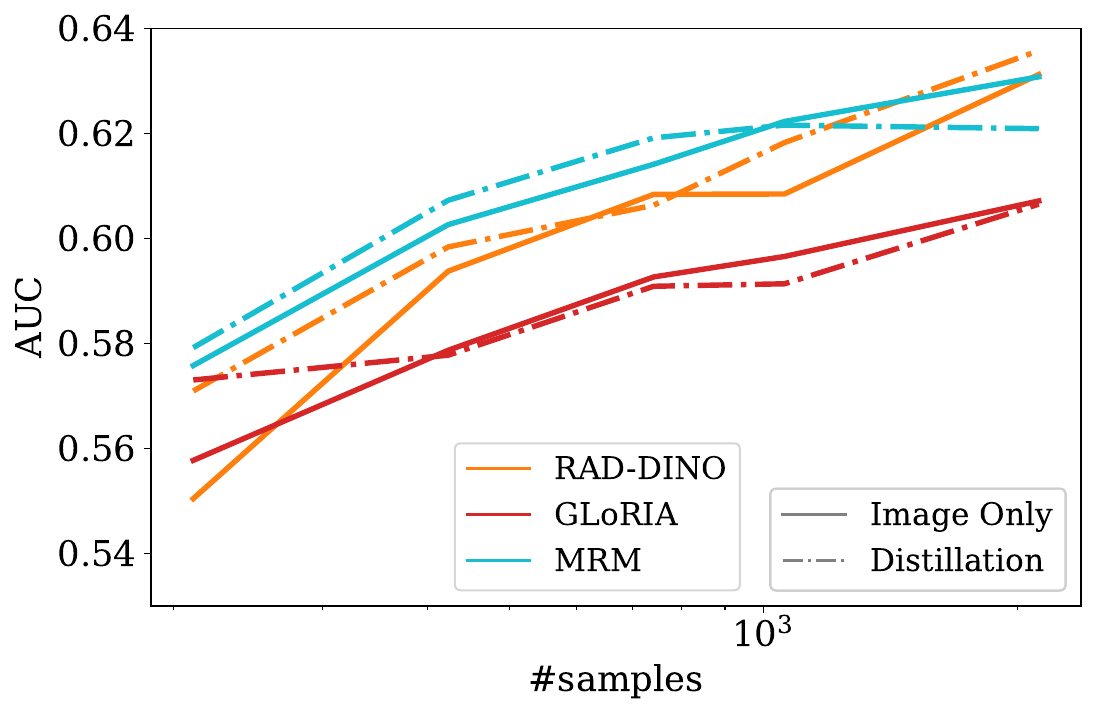}
        \caption{Seed 2}
    \end{subfigure}\hfill
    \begin{subfigure}{0.46\textwidth}
        \centering
        \includegraphics[width=\textwidth]{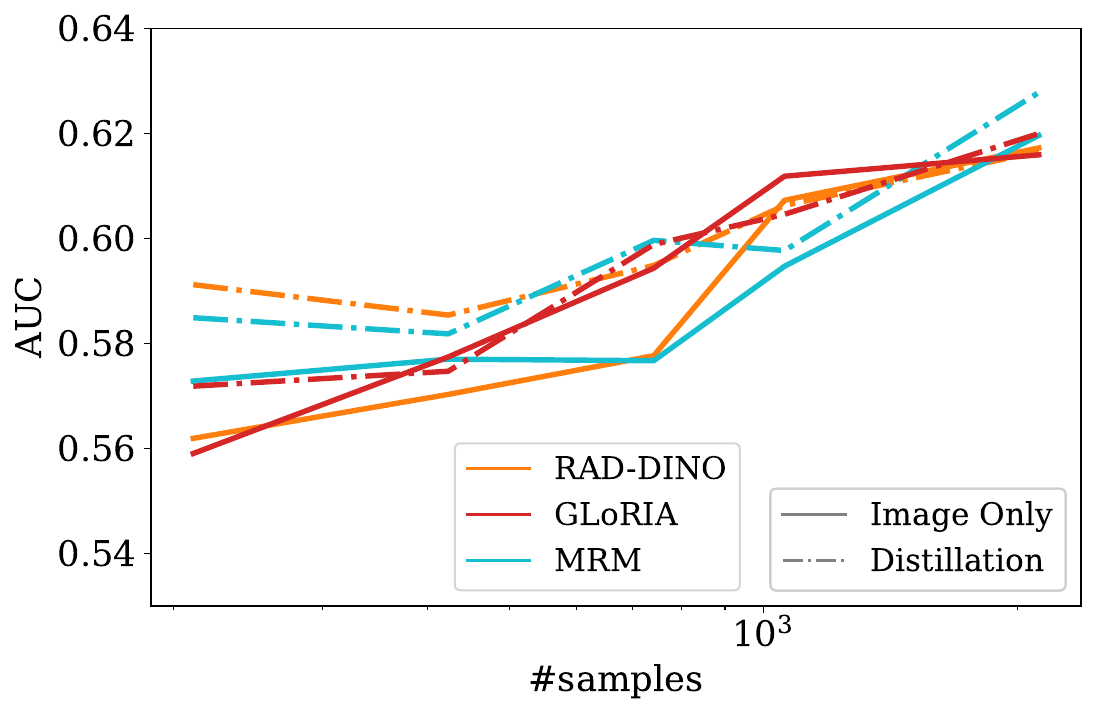}
        \caption{Seed 3}
    \end{subfigure}
    \centering
    \begin{subfigure}{0.46\textwidth}
        \centering
        \includegraphics[width=\textwidth]{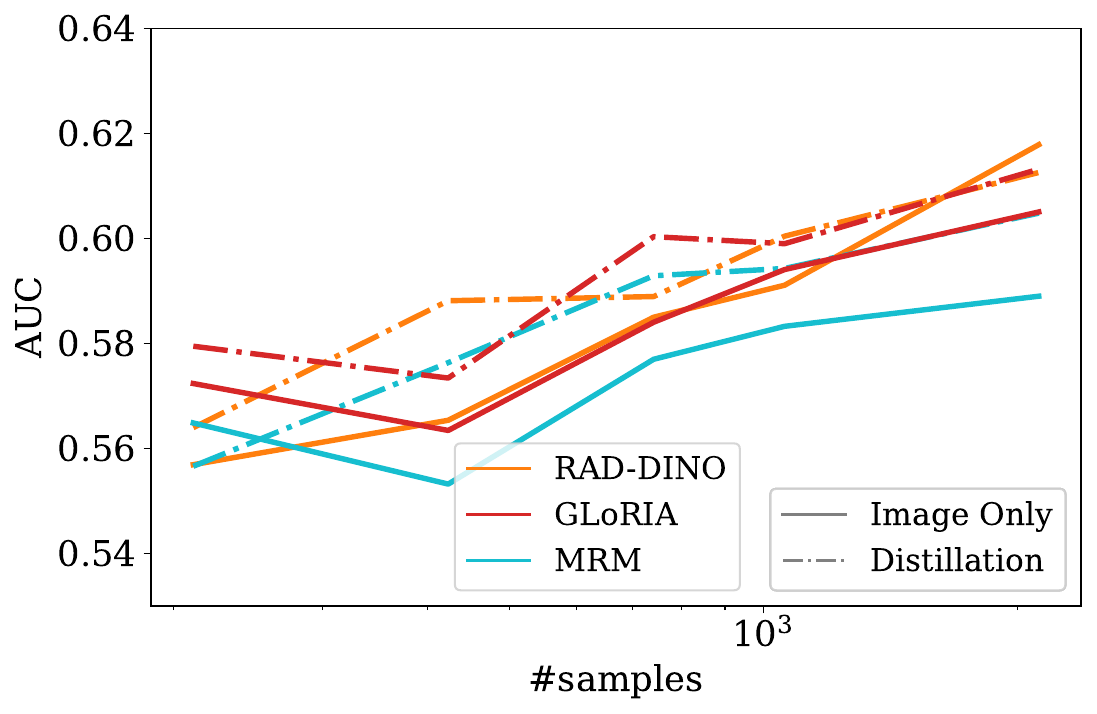}
        \caption{Seed 4}
    \end{subfigure}
    \caption{\textbf{Readmission} - A comparison of image-only models trained with and without distillation across the 5 seeds averaged in Figures \ref{fig:readm_0.05} and \ref{fig:readm_0.5}.}
    \label{fig:readmission_seeds}
\end{figure}

\newpage

\begin{figure}[H]
    \centering
    \begin{subfigure}{0.46\textwidth}
        \centering
        \includegraphics[width=\textwidth]{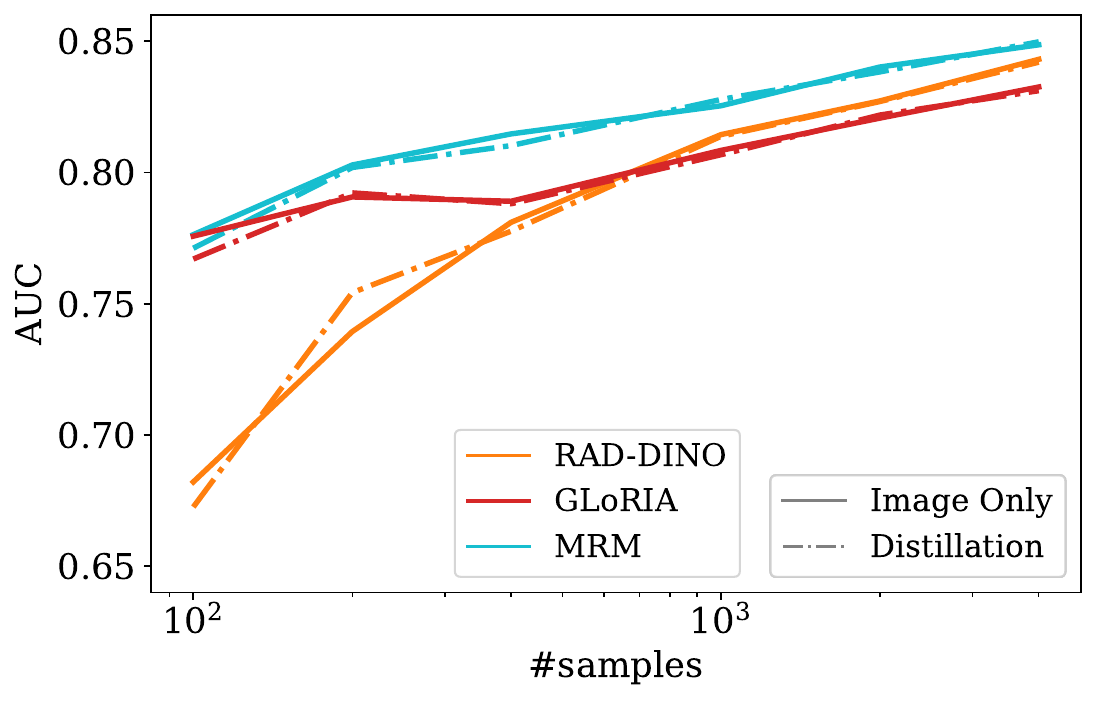}
        \caption{Seed 0}
    \end{subfigure}\hfill
    \begin{subfigure}{0.46\textwidth}
        \centering
        \includegraphics[width=\textwidth]{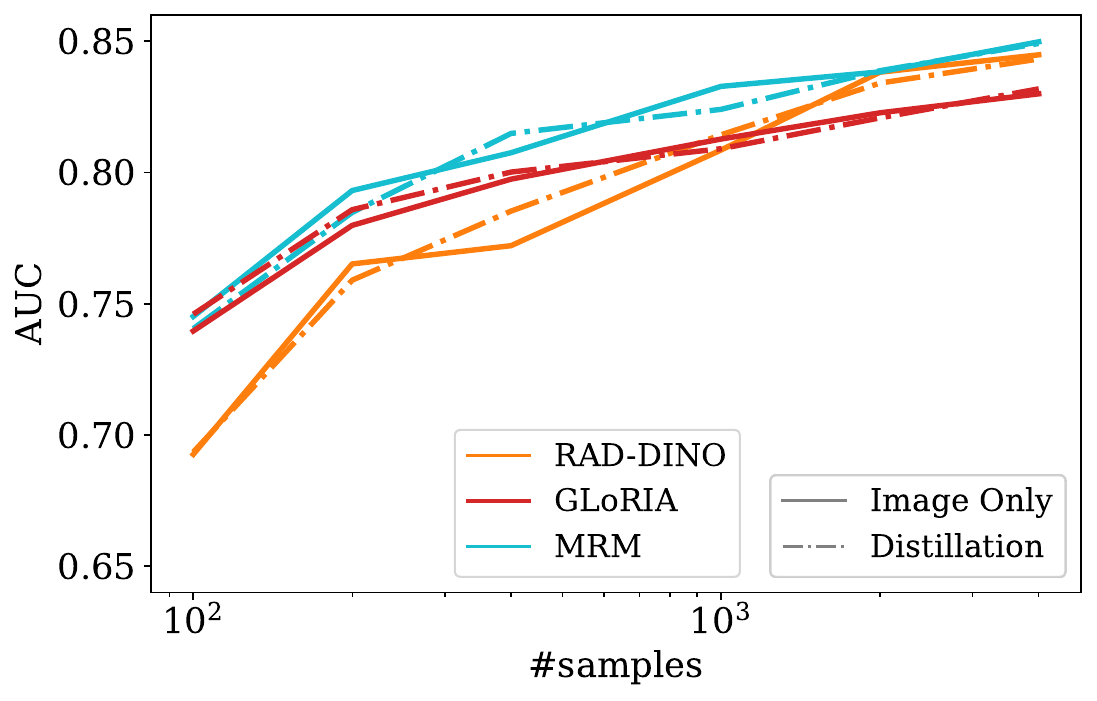}
        \caption{Seed 1}
    \end{subfigure}
    \centering
    \begin{subfigure}{0.46\textwidth}
        \centering
        \includegraphics[width=\textwidth]{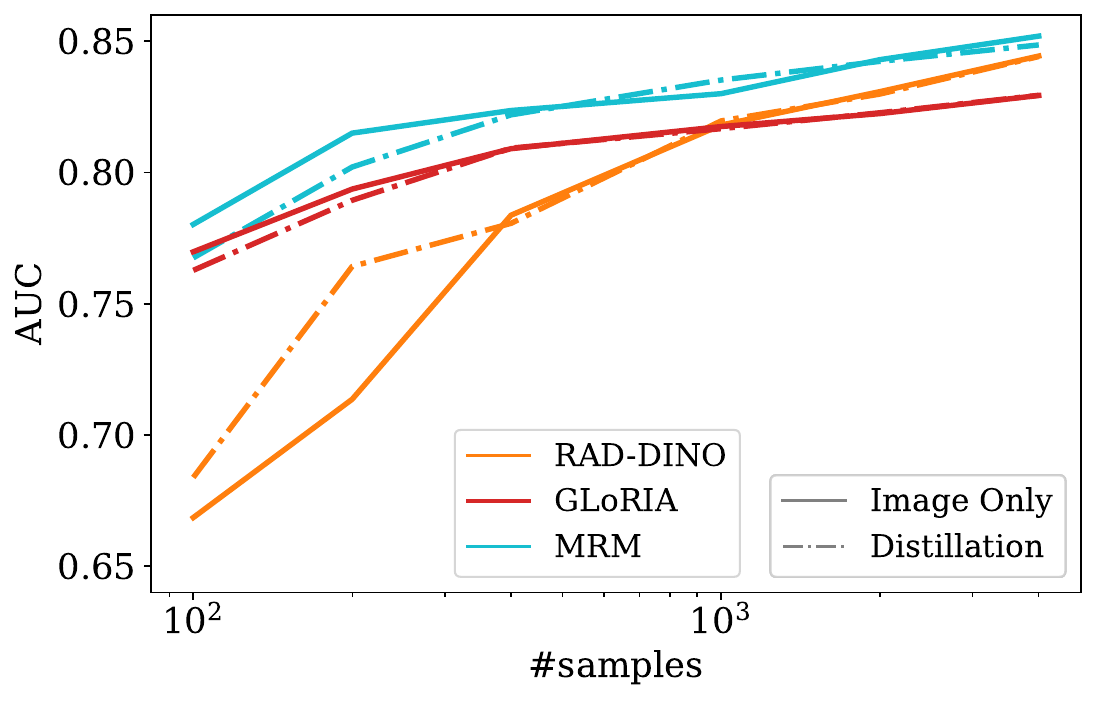}
        \caption{Seed 2}
    \end{subfigure}\hfill
    \begin{subfigure}{0.46\textwidth}
        \centering
        \includegraphics[width=\textwidth]{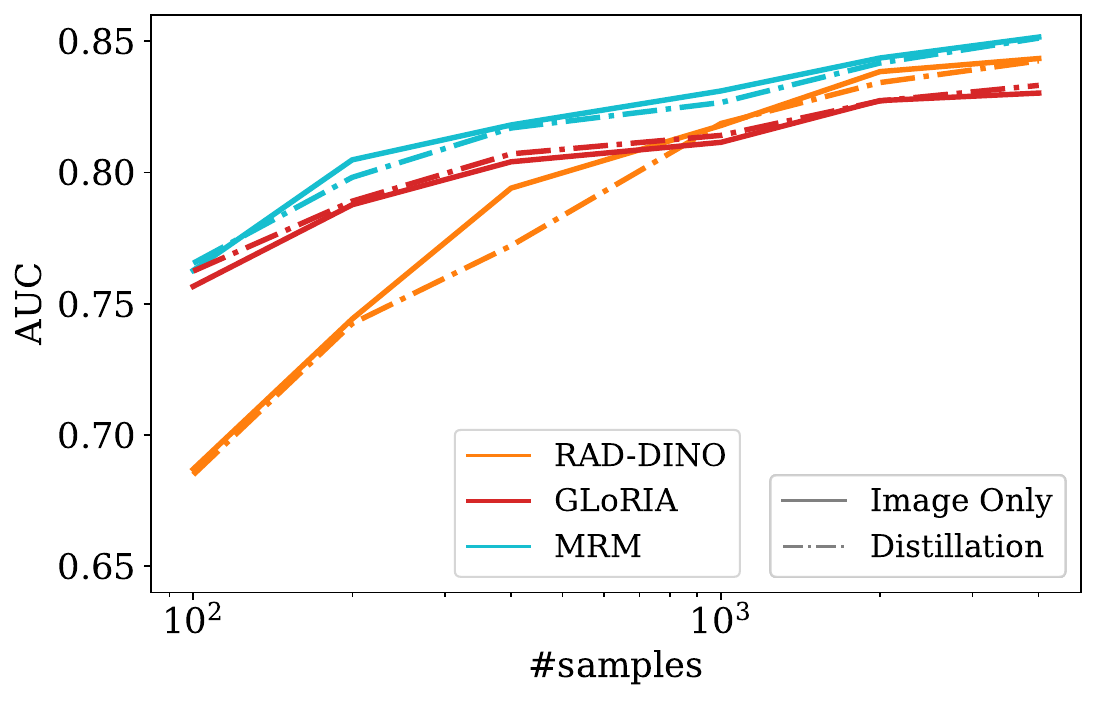}
        \caption{Seed 3}
    \end{subfigure}
    \centering
    \begin{subfigure}{0.46\textwidth}
        \centering
        \includegraphics[width=\textwidth]{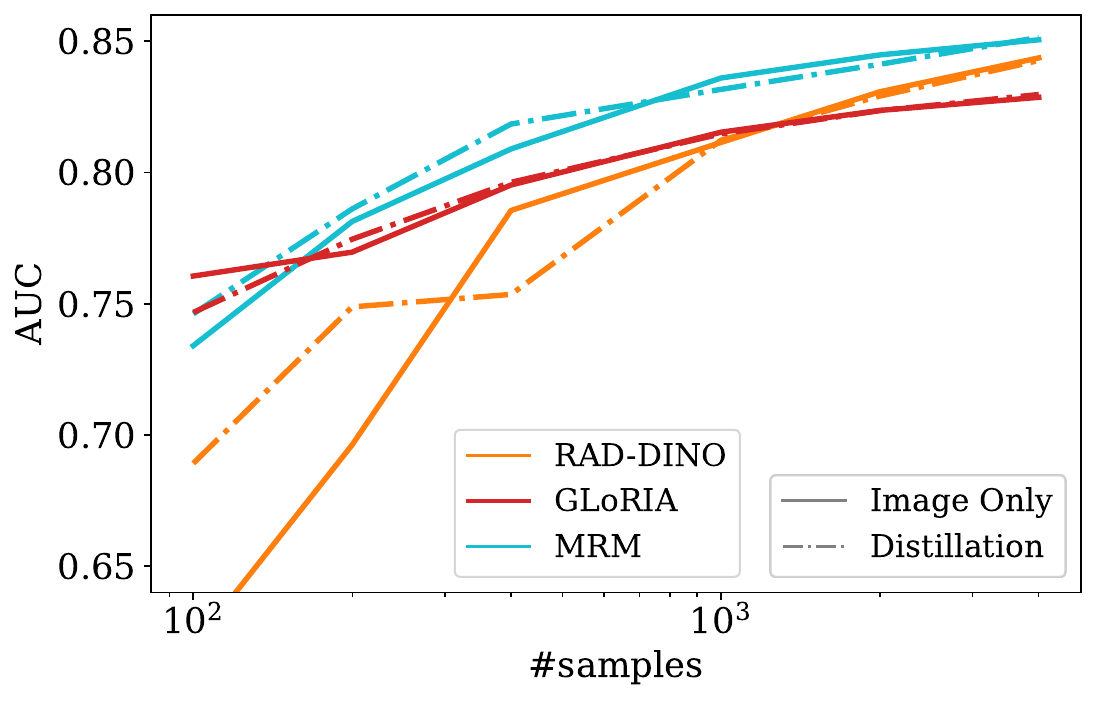}
        \caption{Seed 4}
    \end{subfigure}
    \caption{\textbf{MIMIC 5x1200} - A comparison of image-only models trained with and without distillation across the 5 seeds averaged in Figures \ref{fig:report_labels_self_dist} and \ref{fig:report_0.5}.}
    \label{fig:report_labels_seeds}
\end{figure}

\newpage

\section{Ablation experiments}
\label{sec:ablations}
\subsection{Impact of Attention Head}

\begin{figure}[H]
\centering
    \begin{subfigure}[T]{0.44\textwidth}
       \centering
        \includegraphics[width=0.95\linewidth]{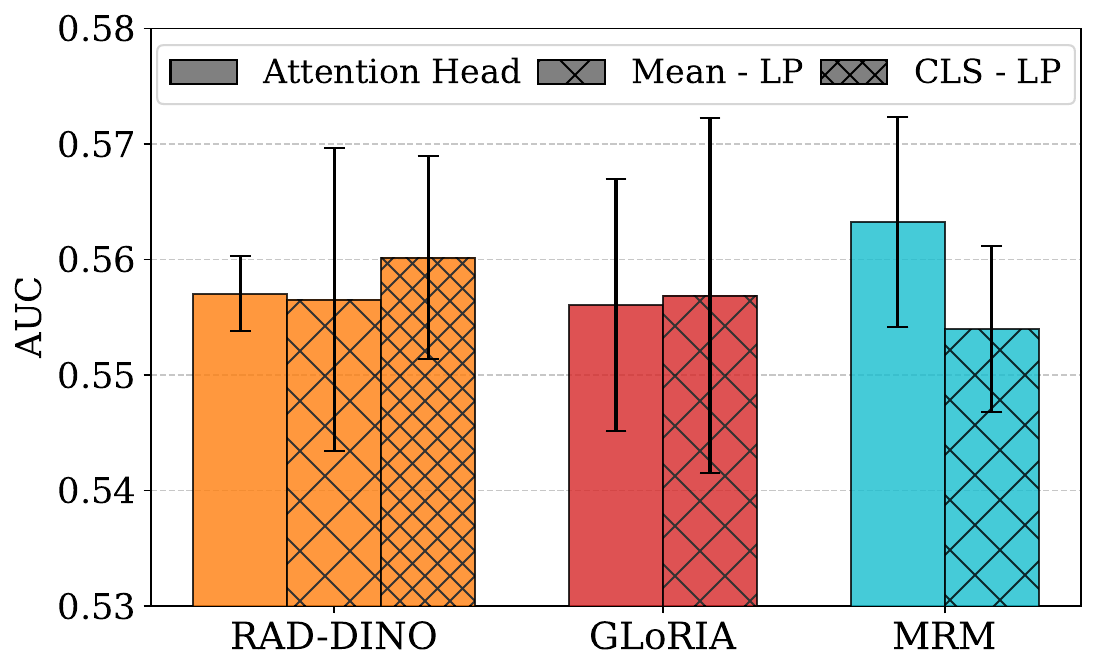}
        \caption{$5 \%$ of training data (211 samples)}
    \end{subfigure}
    \begin{subfigure}[T]{0.44\textwidth}
        \centering
        \includegraphics[width=0.95\linewidth]{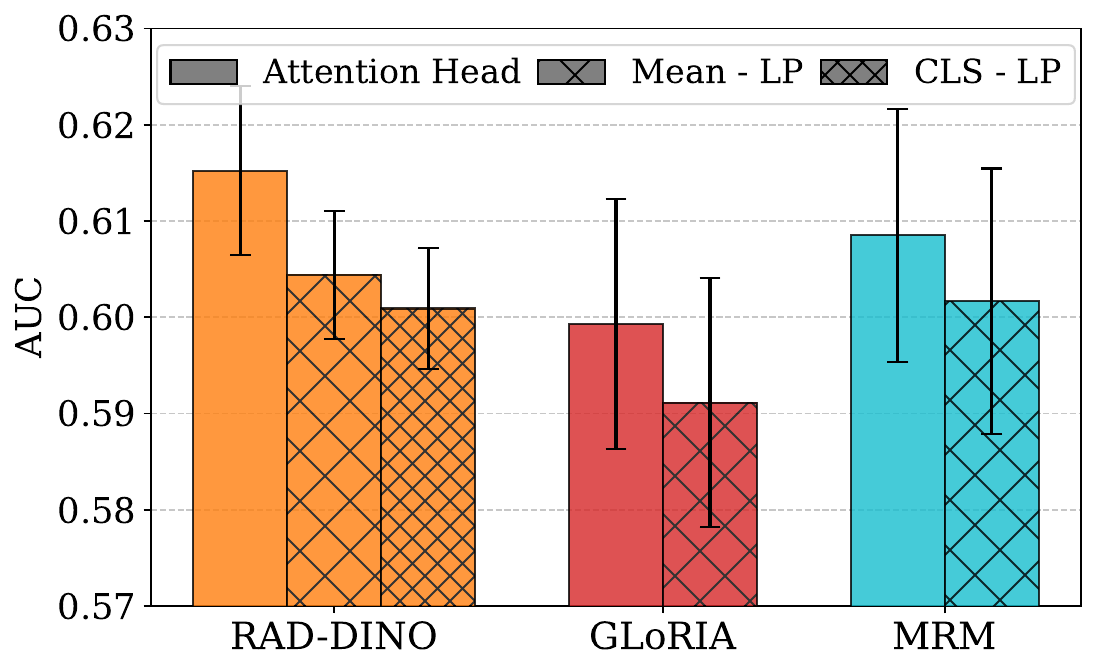}
        \caption{$50 \%$ of training data (2119 samples)}
    \end{subfigure}
    \caption{\textbf{Readmission} - Ablation of different fine-tuning heads on the image-only model. "Attention Head" is the self-attention head used in all experiments in the main paper, "Mean - LP" is mean-pooling followed by linear probing, and "CLS - LP" uses linear probing on the CLS token embedding.}
    \label{fig:readmission_lp}
\end{figure}

\begin{figure}[H]
\centering
    \begin{subfigure}[T]{0.44\textwidth}
       \centering
        \includegraphics[width=0.95\linewidth]{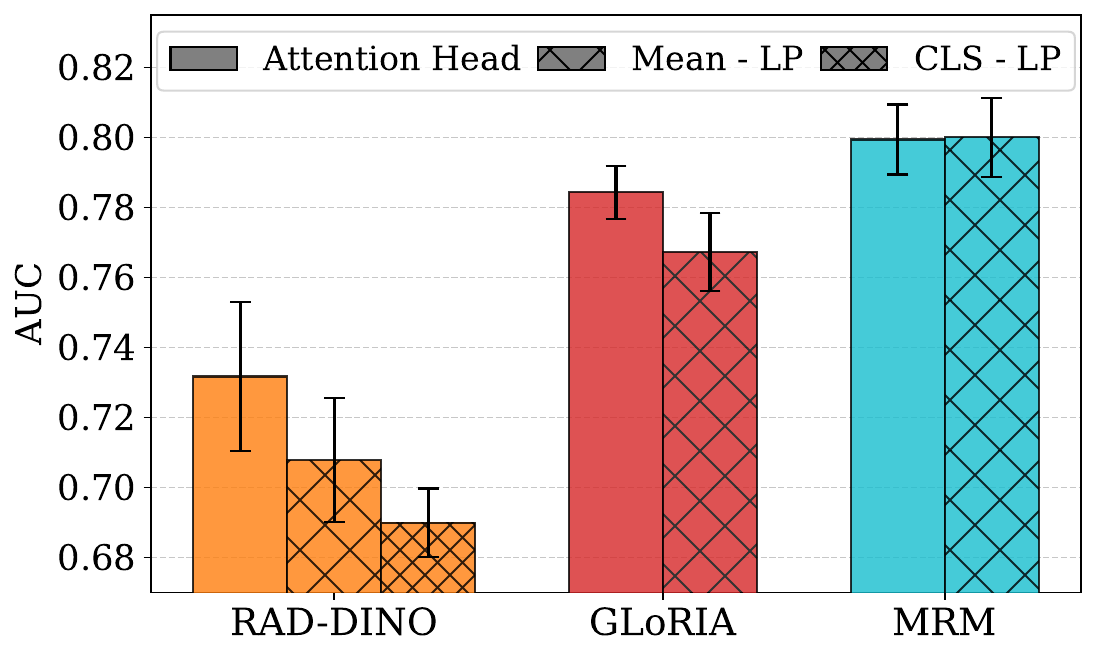}
        \caption{$5 \%$ of training data (200 samples)}
        \label{fig:report_labels_lp_0.05}
    \end{subfigure}
    \begin{subfigure}[T]{0.44\textwidth}
        \centering
        \includegraphics[width=0.95\linewidth]{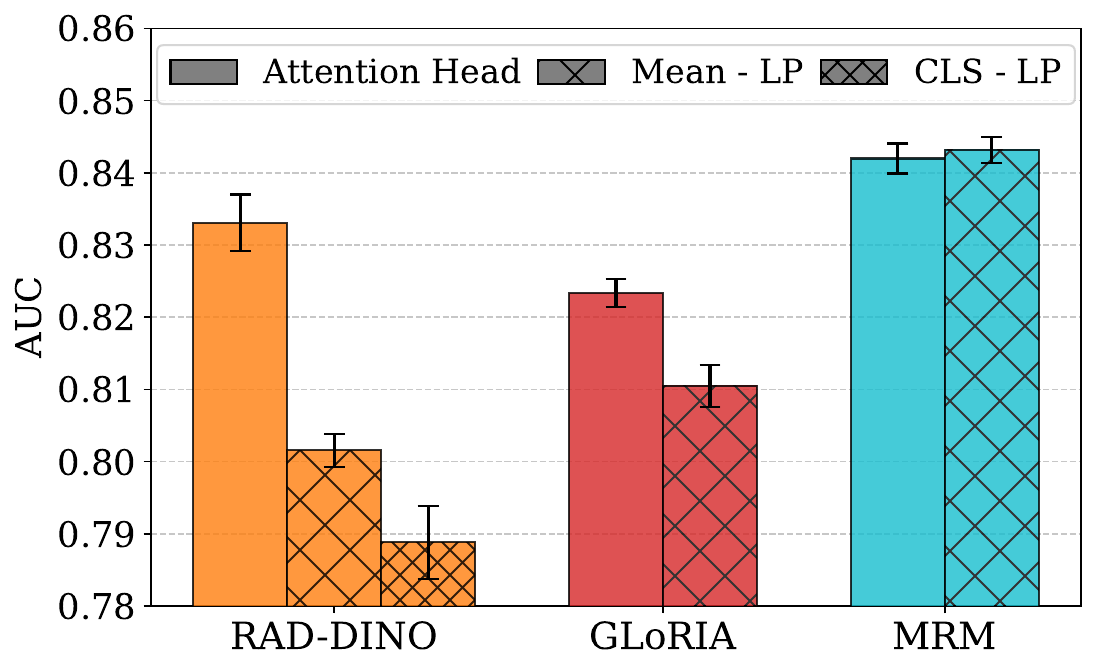}
        \caption{$50 \%$ of training data (2000 samples)}
        \label{fig:report_labels_lp_0.5}
    \end{subfigure}
    \caption{\textbf{MIMIC 5x1200} - Ablation of different fine-tuning heads on the image-only model. "Attention Head" is the self-attention head used in all experiments in the main paper, "Mean - LP" is mean-pooling followed by linear probing, and "CLS - LP" uses linear probing on the CLS token embedding.}
    \label{fig:report_labels_lp}
\end{figure}

Figures \ref{fig:readmission_lp} and \ref{fig:report_labels_lp} demonstrate the impact of the fine-tuning head described in Section \ref{sec:3.2}. ''Attention Head'' corresponds to the self-attention head used in all experiments in the main paper. For ''Mean - LP'', we have instead applied mean pooling over all local embeddings followed by linear probing. Lastly, we evaluated the performance using the RAD-DINO CLS token embedding instead of mean-pooling ("CLS - LP"). We used a learning rate of $10^{-3}$ when performing linear probing (mean- and cls-based). In these experiments, the self-attention head performs noticeably better as the number of samples increases, especially in the case of RAD-DINO. Furthermore, Figure \ref{fig:report_labels_lp} highlights the limitations of only using the CLS token.

\subsection{Temperature parameter}

\begin{figure}[H]
\centering
    \begin{subfigure}[T]{0.44\textwidth}
       \centering
        \includegraphics[width=0.95\linewidth]{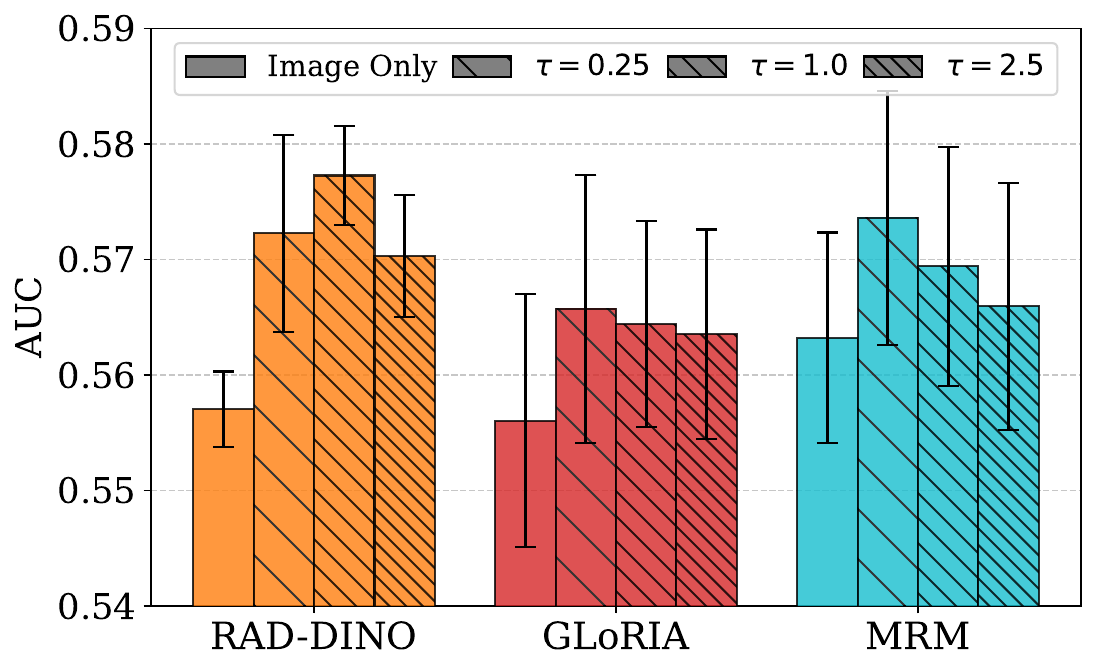}
        \caption{$5 \%$ of training data (211 samples)}
    \end{subfigure}
    \begin{subfigure}[T]{0.44\textwidth}
        \centering
        \includegraphics[width=0.95\linewidth]{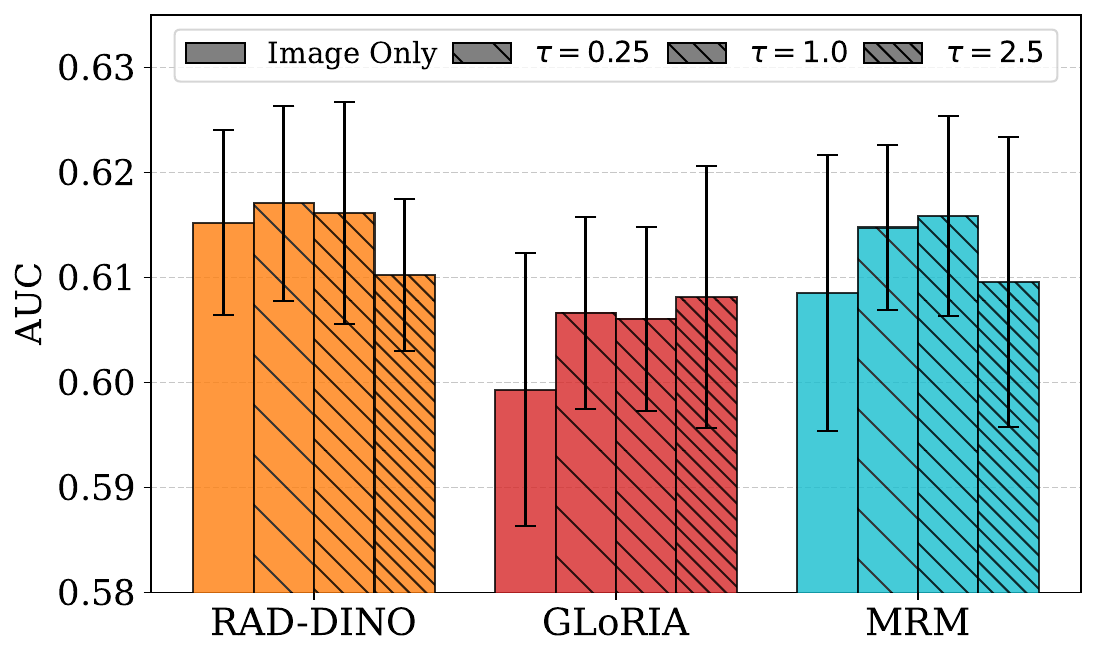}
        \caption{$50 \%$ of training data (2119 samples)}
    \end{subfigure}
    \caption{\textbf{Readmission} - Distillation performance with different temperatures $\tau$.}
    \label{fig:readmission_temp}
\end{figure}

\begin{figure}[H]
\centering
    \begin{subfigure}[T]{0.44\textwidth}
       \centering
        \includegraphics[width=0.95\linewidth]{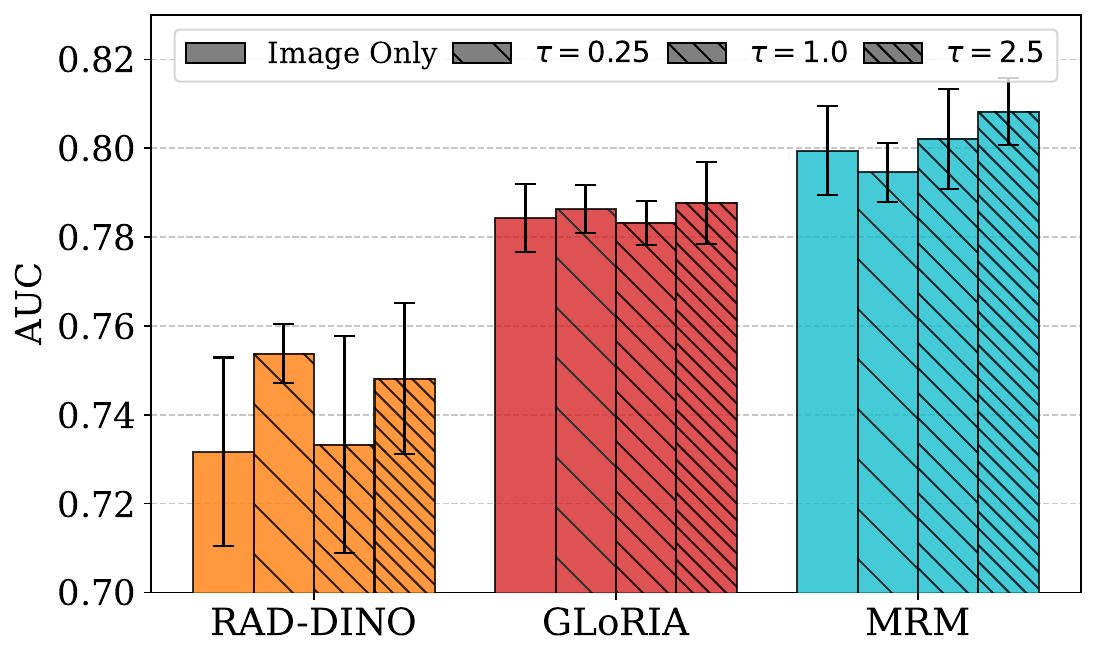}
        \caption{$5 \%$ of training data (200 samples)}
        \label{fig:report_labels_temp_0.05}
    \end{subfigure}
    \begin{subfigure}[T]{0.44\textwidth}
        \centering
        \includegraphics[width=0.95\linewidth]{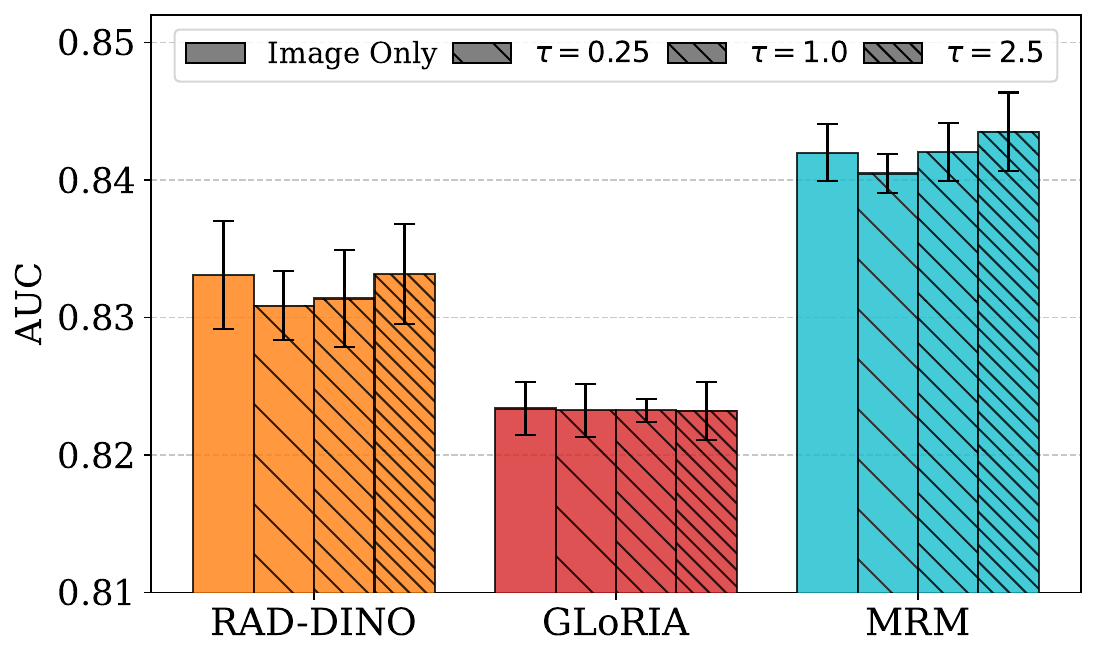}
        \caption{$50 \%$ of training data (2000 samples)}
    \end{subfigure}
    \caption{\textbf{MIMIC 5x1200} - Distillation performance with different temperatures $\tau$.}
    \label{fig:report_labels_temp}
\end{figure}

We perform an ablation experiment on the temperature parameter ($\tau$) in Equation \ref{eq:2} on \textbf{Readmission} (Figure \ref{fig:readmission_temp}) and \textbf{MIMIC 5x1200} (Figure \ref{fig:report_labels_temp}). The distillation models consistently outperform the image-only baseline, regardless of temperature, on the \textbf{Readmission} dataset. On \textbf{MIMIC 5x1200}, the choice of $\tau$ seems to have a modest impact (except for RAD-DINO in Figure \ref{fig:report_labels_temp_0.05}).

\subsection{Choice of text model}

\begin{figure}[H]
    \centering
    \includegraphics[width=0.95\textwidth]{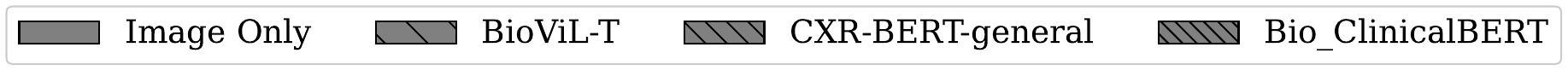}

    \centering
    \begin{subfigure}[T]{0.44\textwidth}
       \centering
        \includegraphics[width=0.95\linewidth]{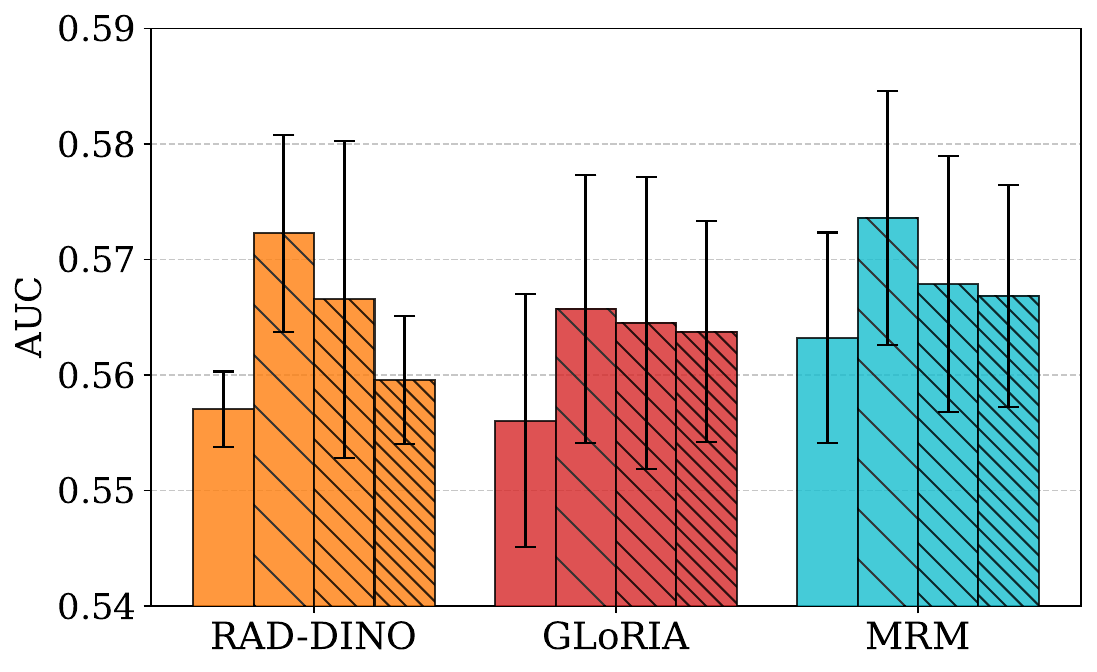}
        \caption{$5 \%$ of training data (211 samples)}
    \end{subfigure}
    \begin{subfigure}[T]{0.44\textwidth}
        \centering
        \includegraphics[width=0.95\linewidth]{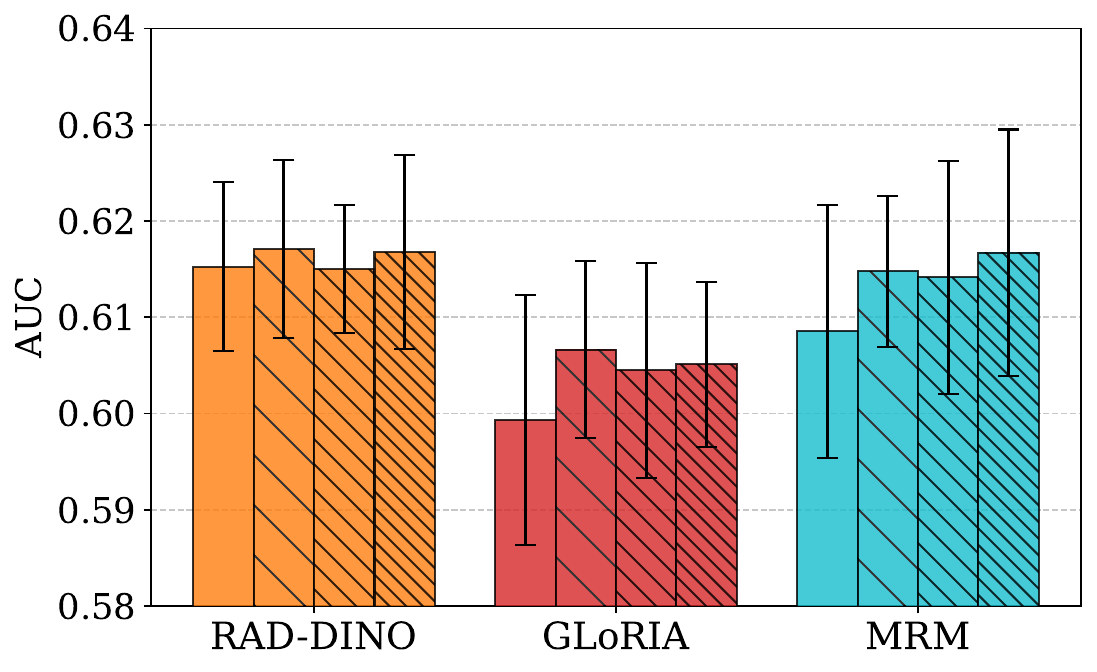}
        \caption{$50 \%$ of training data (2119 samples)}
    \end{subfigure}
    \caption{\textbf{Readmission} - Distillation from teachers trained with different text backbones.}
    \label{fig:readmission_text}
\end{figure}

\begin{figure}[H]
    \centering
    \includegraphics[width=0.95\textwidth]{plots/text_ablation/text_legend_only.pdf}

\centering
    \begin{subfigure}[T]{0.44\textwidth}
       \centering
        \includegraphics[width=0.95\linewidth]{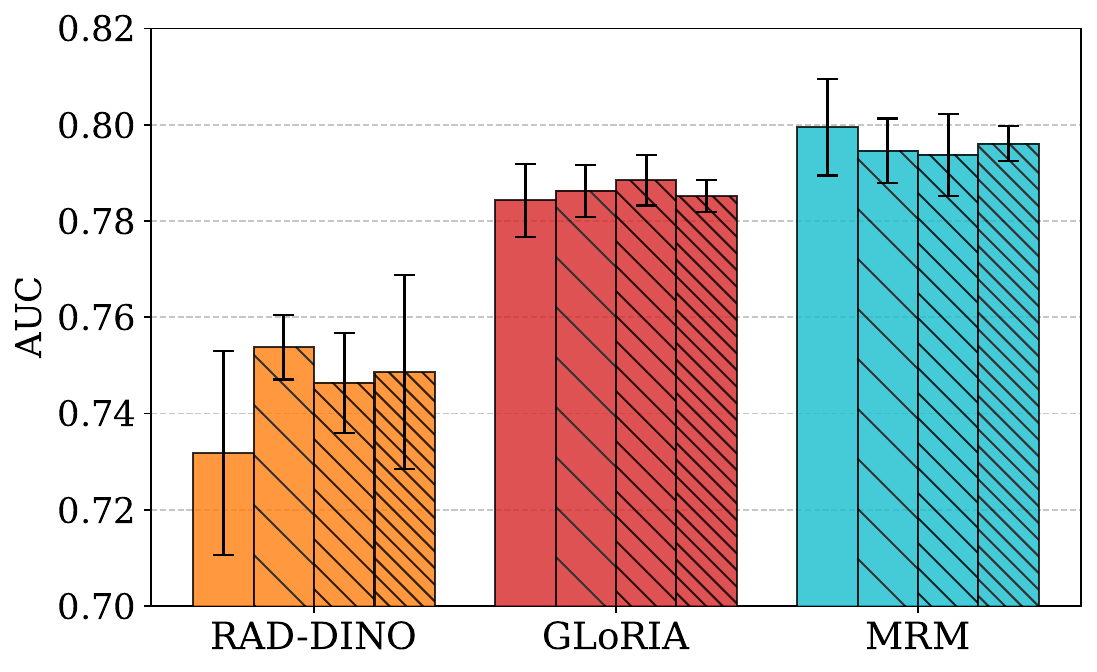}
        \caption{$5 \%$ of training data (200 samples)}
        \label{fig:report_labels_text_0.05}
    \end{subfigure}
    \begin{subfigure}[T]{0.44\textwidth}
        \centering
        \includegraphics[width=0.95\linewidth]{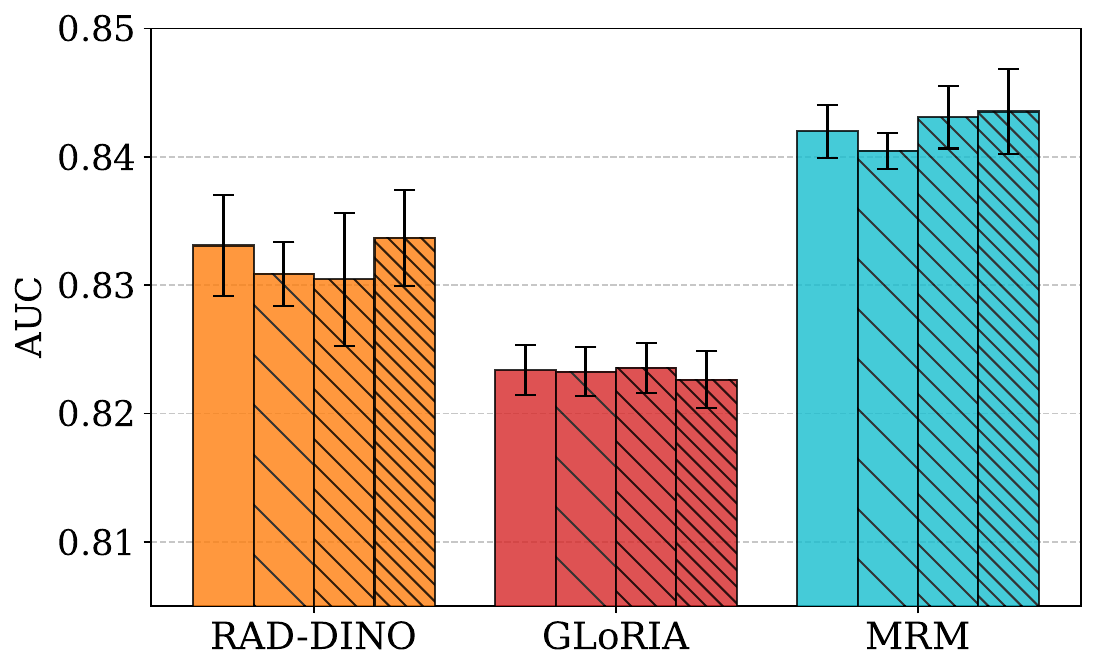}
        \caption{$50 \%$ of training data (2000 samples)}
    \end{subfigure}
    \caption{\textbf{MIMIC 5x1200} - Distillation from teachers trained with different text backbones.}
    \label{fig:report_labels_text}
\end{figure}

To evaluate the impact of the text model in our distillation setup, we train two additional teacher models that each use a different BERT encoder. Apart from the BioViL-T, we use CXR-BERT-general \citep{boecking2022making}, trained with radiology reports, but without the image-based fine-tuning of BioViL-T. We further include Bio\_ClinicalBERT \citep{alsentzer2019publicly}, which has been trained on clinical notes from MIMIC-III, but not radiology reports. Figures \ref{fig:readmission_text} and \ref{fig:report_labels_text} show the performance of the image-only students distilled from these teachers. The results demonstrate that while the choice of text model impacts distillation quality, benefits are observed for all three on the \textbf{Readmission} dataset. While it would be interesting to explore using the accompanying text encoders for the VLM backbones, not all of these are available, and we limit ourselves to the three covered here to make the comparison as fair as possible. CXR-BERT-general and Bio\_ClinicalBERT are available on \href{https://huggingface.co/microsoft/BiomedVLP-CXR-BERT-general}{https://huggingface.co/microsoft/BiomedVLP-CXR-BERT-general} and \href{https://huggingface.co/emilyalsentzer/Bio_ClinicalBERT}{https://huggingface.co/emilyalsentzer/Bio\_ClinicalBERT} respectively.

\section{Compute}
\label{sec:compute}

Experiments were primarily performed on the NVIDIA T4 (16GB) GPU. RAD-DINO had to be run on an NVIDIA A100 (40GB) to avoid memory constraints. No experiment ran for more than 8 hours, with the majority completing in less than half that time. Given this, we approximate that all experiments run in Figure \ref{fig:backbone_comp} require less than 1,000 hours. Training of the additional teacher, distillation, and self-distillation models in Figure \ref{fig:barplot} meant that we had 4 methods for each of the 6 backbones over 5 seeds. Again using the upper limit of 8 hours per run, each subfigure in \ref{fig:barplot} took less than $8 \cdot 6 \cdot 4 \cdot 5 = 960$ hours to run.

\section{Dataset Licenses}
\label{sec:license}

\paragraph{MIMIC} JPEG images and labels extracted from reports were collected from MIMIC-CXR-JPG 2.1.0 \href{https://physionet.org/content/mimic-cxr-jpg/2.1.0/}{https://physionet.org/content/mimic-cxr-jpg/2.1.0/}. Free-text reports and image metadata were fetched from MIMIC-CXR 2.1.0 \href{https://physionet.org/content/mimic-cxr/2.1.0/}{https://physionet.org/content/mimic-cxr/2.1.0/}. Patient admission information (used for the \textbf{3-Day Discharge} dataset) was gathered from MIMIC-IV 3.1 \href{https://physionet.org/content/mimiciv/3.1/}{https://physionet.org/content/mimiciv/3.1/}. All data is provided under the PhysioNet Credentialed Health Data License 1.5.0 \href{https://physionet.org/content/mimic-cxr/view-license/2.1.0/}{https://physionet.org/content/mimic-cxr/view-license/2.1.0/}.

\paragraph{INSPECT} The INSPECT dataset was downloaded from \href{https://stanfordaimi.azurewebsites.net/datasets/318f3464-c4b6-4006-9856-6f48ba40ad67}{https://stanfordaimi.azurewebsites.net/datasets/318f3464-c4b6-4006-9856-6f48ba40ad67}. The data is licensed under the Stanford University Dataset Research Use Agreement, provided on the same page.


\end{document}